\documentclass[journal]{IEEEtran}

\usepackage{amssymb,amsmath,amsthm}
\usepackage{color}
\usepackage{cite}

\usepackage{algorithm,url}
\usepackage[noend]{algpseudocode}

\usepackage{calc}
\usepackage{amsmath}
\usepackage{xcolor}
\usepackage{subcaption} 
\usepackage[font=footnotesize]{caption}
\usepackage{graphicx}

\DeclareMathOperator*{\argmin}{argmin} 

\newtheorem{theorem}{Theorem}
\newtheorem{corollary}{Corollary}

\usepackage[inline, shortlabels]{enumitem}

\usepackage{verbatim}  

\newcommand{\Pro}{\mathsf{P}}
\newcommand{\Exp}{\mathsf{E}}
\newcommand{\cX}{\mathcal{X}}
\newcommand{\cA}{\mathcal{A}}
\newcommand{\cG}{\mathcal{G}}
\newcommand{\cE}{\mathcal{E}}

\newcommand{\td}{\text{d}}
\newcommand{\bR}{\mathbb{R}}

\newcommand{\cS}{\mathcal{S}}
\newcommand{\cQ}{\mathcal{Q}}

\renewcommand{\vec}[1]{\boldsymbol{#1}}
\newcommand{\vx}{\vec{x}}
\newcommand{\vy}{\vec{y}}

\newcommand{\be}{\begin{equation}}
\newcommand{\ee}{\end{equation}}
\newcommand{\ignore}[1]{{}}

\begin{document}

\title{Online Multivariate Anomaly Detection and Localization for High-dimensional Settings}

\author{Mahsa Mozaffari,~Yasin Yilmaz
\thanks{Authors are with the Department of Electrical Engineering, University of South Florida, Tampa, FL 33620, USA (e-mail: mmozaffari@mail.usf.edu, yasiny@usf.edu).}
}

\maketitle

\begin{abstract}
This paper considers the real-time detection of anomalies in high-dimensional systems. The goal is to detect anomalies quickly and accurately so that the appropriate countermeasures could be taken in time, before the system possibly gets harmed. We propose a sequential and multivariate anomaly detection method that scales well to high-dimensional datasets. The proposed method follows a nonparametric, i.e., data-driven, and semi-supervised approach, i.e., trains only on nominal data. Thus, it is applicable to a wide range of applications and data types. Thanks to its multivariate nature, it can quickly and accurately detect challenging anomalies, such as changes in the correlation structure and stealth low-rate cyberattacks. Its asymptotic optimality and computational complexity are comprehensively analyzed. In conjunction with the detection method, an effective technique for localizing the anomalous data dimensions is also proposed. We further extend the proposed detection and localization methods to a supervised setup where an additional anomaly dataset is available, and combine the proposed semi-supervised and supervised algorithms to obtain an online learning algorithm under the semi-supervised framework. The practical use of proposed algorithms are demonstrated in DDoS attack mitigation, and their performances are evaluated using a real IoT-botnet dataset and simulations. 
\end{abstract}
\begin{IEEEkeywords}
Anomaly detection, anomaly localization, online learning, nonparametric methods, DDoS attack mitigation
\end{IEEEkeywords}
\IEEEpeerreviewmaketitle
\section{Introduction}

Anomaly detection is an important problem dealing with the detection of abnormal data patterns \cite{SURVEY}. It has applications in a variety of different domains, such as cybersecurity \cite{newinfobased}, medical health care\cite{medicAD}, quality control, etc. The importance of anomaly detection lies in the fact that an anomaly in the observation data may be a sign of an unwanted event such as failure, malicious activity, etc. in the underlying system. Therefore, accurate detection of such data patterns will allow proper countermeasures to be taken by the domain specialist to counteract any possible harm. To name a few examples, an anomaly in the MRI image could be due to the presence of a malignant tumor in the brain; and anomalous observations in the network traffic data could mean that the network is under a cyber-attack. 

The advances in various technologies such as Internet-of-Things (IoT) devices and sensors, and wireless communications, have enabled the real-time monitoring of systems for detecting events of interest. In many modern and complex systems such as IoT networks, network-wide traffic monitoring systems, environmental monitoring systems, etc. massive amounts of heterogeneous data are generated, which require real-time processing for timely detection of anomalous events.
As an example, automated vehicles or advanced driver-assistance systems today are equipped with modules comprising a large number of sensors and actuators for control and safety purposes. Due to the catastrophic consequences of any fault in perceiving the environment or failure in a component of the system, as well as being compromised by hackers, it is crucial to preserve the robustness of the vehicle. To this end, the high-dimensional measurements from sensors need to be monitored and analyzed in real-time to detect anomalies such as sudden increase of speed, abnormal petrol consumption, anomalies in radar sensors and camera sensing, etc. \cite{BIGDATASURVEY}. Accurate and light-weight anomaly detection methods that can scale well to large systems are needed to be able to address such big data challenges in real-time.

Anomaly detection methods on univariate data streams have been studied thoroughly in the literature. However, little work has been done on multivariate anomaly detection, which has the potential to achieve quicker and more accurate detection than univariate anomaly detection by capturing more anomaly evidence in the interactions between system dimensions. Statistical approaches to anomaly detection assume anomaly to be a change in the probability distribution of the observations, such as a change in the mean, variance, or correlation structure between the data-streams. One important application for detecting changes in the correlation structures is finance, where the correlation structures between high-dimensional processes modeling the exchange rates and market indexes are important for the right choice of asset allocation in portfolio \cite{CORR}. Furthermore, in social networks, it is important to detect abrupt changes in interactions between the nodes; and in communication networks, it is of interest to detect highly correlated traffic in a network \cite{taposh1}. Distributed Denial of Service (DDoS) attacks to power grid through synchronous switching on/off of high-wattage IoT devices is another example where anomaly is manifested in correlations \cite{soltan2018blackiot}.
Detection of change in correlation structure requires the joint monitoring and multivariate analysis of the data-streams, which in turn, leads to the high-dimensionality challenge. To overcome this challenge, a desired anomaly detection technique needs to be scalable to high-dimensional data in real time.  

Anomaly detection in many systems such as fraud detection could be the ultimate goal, however, in many scenarios, such as diagnosis systems (e.g., spacecraft monitoring system \cite{spacecraft}) and cybersecurity, it is highly important to provide a degree of interpretation about the detected issue in the system and how to mitigate it. Considering the potential damages caused by failure in mitigation of unexpected behaviors, such as cyber-attacks, detecting anomalies without providing any further information explaining where the anomaly has happened is of limited value to the engineers. 

Motivated by the aforementioned challenges, we investigate an \emph{online multivariate anomaly detection and localization} technique which is simple enough to handle high-dimensional and heterogeneous data in real-time.

\subsection{Related Works}

The problem of anomaly detection, also known as outlier detection or novelty detection, has been an important subject of study in several research communities such as statistics, signal processing, machine learning, information theory, data mining, etc. either specifically for an application domain or as a generic method. To name a few, an SVM classification approach for anomaly detection was proposed in \cite{svmbased}; In \cite{c3,c4,j1} the authors study the outlier detection problem for high-dimensional data that adhere to non-linear manifold structures, which generalizes the linear setting of the well-established Robust PCA problem. Via a new notion of conformity, \cite{c3,j1} present remarkably fast methods which overcome the limitations of the previous work on challenging settings including large number of outliers, coherent outliers and noise contamination. In \cite{c4}, anomaly detection on low-dimensional manifolds is considered, and a novel approach characterizing the samples' ability to represent the data is developed, to identify outlier data; Information theoretic measures were proposed in \cite{infobased} for the intrusion detection problem; and two new information metrics for DDoS attack detection was introduced in \cite{newinfobased}. Due to the challenging nature of the problem and considering the challenges posed by today's technological advances such as big data problems, there is still a need for reconsidering the anomaly detection problem. 

Sequential anomaly detection techniques, compared to the outlier detection techniques \cite{SURVEY}, take also the history of observations into account rather than only the new observations. Sequential techniques are more suitable for real-time systems where timely and accurate detection of anomalies is important. The Cumulative Sum (CUSUM) detector \cite{CUSUM} is a well-known sequential change detection technique that assumes probabilistic models for nominal and anomalous data points, and computes the cumulative log-likelihood-ratio (LLR) over time, declaring anomaly if the statistic exceeds a predefined threshold. The accuracy of assumed models as well as the estimated parameters are the key factors in the performance of CUSUM and more generally parametric methods. CUSUM is minimax optimum under the condition that the probability distributions before and after the change are completely known \cite{Moustakides}. However, in many real-world applications having \emph{a priori} knowledge about the underlying distributions is not possible. Estimating the probability distributions quickly becomes intractable for high-dimensional data, which includes many unknowns such as the anomaly onset time, the subset of anomalous dimensions, etc., in addition to the parameters of the nominal and anomalous models. To tackle with this complexity, \cite{MEI} proposed a relaxed version of CUSUM in which each data stream is assumed to be independent of others. However, this univariate method is not suitable for detecting changes in the correlation between data streams.
A sequential test for detecting changes in the correlation between variables, as well as localizing the highly correlated variables, in high-dimensional data streams has been proposed in \cite{taposh_hub}. This is a parametric method based on the assumption that the observed vectors are multivariate Gaussian distributed. It is proposed solely for the detection of correlation change between data-streams and does not generalize to other changes in the distribution. In this paper, we are interested in detecting general changes in unknown distributions, including the changes in correlation structure.

$k$-nearest-neighbor ($k$NN) distance-based methods are geometric methods that are based on the assumption that anomalous data instances occur far from the nominal instances. For instance, \cite{GEM} and \cite{GEM-2} have proposed nonparametric outlier detection techniques based on the minimum volume set (MVS) of the nominal data. MVS corresponds to the region of greatest probability density with minimum data volume and is known to be useful for anomaly detection \cite{LMVS} based on the assumption that anomalies occur in the less concentrated regions of the nominal dataset. These nonparametric outlier detection methods estimate the MVS of nominal training samples using $k$NN graphs, and declare a data point as anomalous if it lies outside the MVS. Despite being scalable to high-dimensional and heterogeneous data, they do not consider the temporal anomaly information, and thus are prone to higher false alarm rates compared to sequential anomaly detection methods. 
Similarly, \cite{zhao} proposed a $k$NN graph-based method that computes an anomaly score for each observation and declares an anomaly by thresholding the score value. 
In this paper, as opposed to the outlier detection methods which treat a single outlier as an anomaly, we consider an anomaly to consist of persistent outliers and investigate sequential and nonparametric detection of such anomalies using the temporal information in data streams. 
Recently, \cite{chen} proposed a nonparametric $k$NN-based sequential anomaly detection method for multivariate observations. This method computes the test statistic based on the number of $k$NN edges at different splitting points within a window and stops the test whenever the test statistics exceed a threshold. Due to its window-based nature this method has inherent limitations in achieving small detection delays. It also recomputes the $k$NN graphs at every time instance and for every splitting point, therefore its computational complexity is note suitable for real-time applications. In another recent work, \cite{zambon} proposed a distance-based and CUSUM-like change detection method for attributed graphs. Attributed graphs are first mapped into numeric vectors, and then the distance between the mean response of an observation window and the mean response of the training data are computed via a CUSUM-like sequential algorithm. In addition to the limitations arising from the window-based nature of the method, the local relations between samples are disregarded due to considering only the mean response of the training set. As a result, in cases where training data has a multimodal distribution, this method will not be effective. As compared to \cite{zambon}, we take into account the local relations between the data instances.

\subsection{Contributions}
In this paper, aiming at timely and accurate detection of anomalies in high-dimensional systems we propose two variations of a $k$NN-based sequential anomaly detection method, as well as a unified framework that combines the advantages of both methods. In summary, our contributions in this paper are as follows: 

\begin{itemize}
\item A framework for multivariate, data-driven and sequential detection of anomalies in high-dimensional systems is proposed for both semi-supervised and supervised settings depending on the availability of labeled data. Combining the advantages of supervised and semi-supervised settings, we further introduce an online learning scheme which can effectively detect both known and unknown anomaly types by incorporating the newly detected anomalies into the training set. 
\item Asymptotic optimality of the proposed detection methods in the minimax sense is shown, and comprehensive analysis for computational complexity is provided.
\item An anomaly localization technique to identify the problematic data dimensions is also proposed based on the proposed detection methods.
\item The practicality of the proposed anomaly detection and localization methods is demonstrated on mitigating DDoS attacks through simulations and a real dataset. 
\end{itemize}

\subsection{Organization and Notations}

The rest of the paper is organized as follows. In Section~\ref{sec:formulation}, the mathematical formulation of the considered anomaly detection problem and the relevant background information are provided. 
We present the proposed anomaly detection and localization methods in Sections~\ref{sec:proposed} and \ref{sec:localization}.
Section~\ref{sec:ddos} presents the application of our proposed methods in DDoS attack mitigation. 
Finally, we conclude the paper in Section~\ref{sec:conc}.

Vectors and matrices are represented by boldface lowercase and uppercase letters, respectively. Script letters denote sets, e.g., $\mathcal{X}$. Vectors are organized in a column unless otherwise stated. Probability and expectation are denoted with $\Pro$ and $\Exp$, respectively.

\section{Problem Formulation}
\label{sec:formulation}

Suppose that a system is observed through $d$-dimensional observations ${\mathcal{X}_t = \{\vx_1,\vx_2,\ldots,\vx_t \}}$ in time. The objective is to detect an anomaly occurring at an unknown time $\tau$ as soon as possible while satisfying a false alarm constraint.
This problem can be formulated as a change detection problem as follows:
\begin{align} 
\label{eq:hyp}
& f = f_0, \; t<\tau, \quad f = f_1 (\neq f_0), \; t\geq \tau,
\end{align}
where $f$ is the true probability distribution of observations, $f_0$ and $f_1$ are the nominal and anomaly probability distributions, respectively. The objective of the problem is to find the anomaly time $T$ that minimizes the average detection delay while satisfying a false alarm constraint, i.e., 
\be 
\label{eq:obj}
\inf_T \Exp_{\tau}[(T - \tau)^+]  ~~\text{subject to}~~ \Exp_\infty[T] \geq \beta,
\ee
where $E_{\tau}$ represents the expectation given that change occurs at $\tau$, $(.)^+ = \max(., 0)$, and $\Exp_\infty$ denotes the expectation given that no change occurs, i.e., the expectation of false alarm period. 

Lorden's minimax problem is a commonly used version of the above problem \cite{lorden}, in which the goal is to minimize the worst-case average detection delay subject to a false alarm constraint: 
\be
\label{eq:minimax}
\inf_T \sup_{\tau} \text{ess} \sup_{\mathcal{X}_{\tau}} E_{\tau}[(T - \tau)^+|\mathcal{X}_{\tau}]  ~~\text{s.t.}~~ E_\infty[T] \geq \beta,
\ee
where ``ess sup" denotes essential supremum which is equivalent to supremum in practice. In simple words, the minimax criterion minimizes the average detection delay for the least favorable change-point and the least favorable history of measurements up to the change-point while the average false alarm period is lower bounded by $\beta$. 

The CUSUM test provides the optimum solution to the minimax problem \cite{Moustakides}, given by \eqref{eq:minimax}:
\begin{align} 
\label{eq:cusum_proc}
\begin{split}
& S_t = \max\{ 0, S_{t-1}+\ell_t \}, \\
& T_c = \inf\{t: S_t \geq h_c  \},
\end{split}
\end{align}
where $T_c$ is the stopping time, $\ell_t = \log\frac{f_1(\vx_t)}{f_0(\vx_t)}$ is the log-likelihood ratio at time $t$, $S_0=0$, and $h_c$ is a decision threshold, selected in a way to satisfy a given false alarm constraint. 
Considering $\ell_t$ as a statistical evidence for anomaly the CUSUM algorithm continues accumulating it, and declares an anomaly the first time the accumulated evidence $S_t$ exceeds a threshold $h_c$, that is chosen sufficiently large for reliable detection. CUSUM requires the complete knowledge of the probability distributions $f_0$ and $f_1$. However, in real-world applications, the true probability distributions are typically unknown. Even when $f_0$ and $f_1$ are known up to their parameters, and the parameters are estimated using the maximum likelihood approach, the procedure known as Generalized CUSUM (G-CUSUM) achieves only asymptotic optimality. Moreover, CUSUM and in general parametric methods are limited to the detection of certain anomaly types whose true probability distribution matches the assumed $f_1$ well. 

In high-dimensional problems that require multivariate analysis, estimating the nominal probability distribution is typically not tractable, especially when the data dimensions are heterogeneous, e.g., environmental sensor data consisting of wind speed, direction, air temperature, pressure, humidity, weather condition (whether it is rainy, sunny or cloudy), etc. Considering the wide range of possible anomalies it is even more intractable to estimate the anomaly probability distribution. In such problems, knowing the probability distributions and parameters is highly complicated if not impossible, limiting the applicability of CUSUM and parametric methods in general.

\section{Proposed Detection Methods} 
\label{sec:proposed}

We recently proposed a $k$NN-based sequential anomaly detection method called Online Discrepancy Test (ODIT) \cite{odit}, and applied it to cyber-attack detection in smart grid \cite{icmla} and in intelligent transportation systems \cite{itsc}. In this section, we (i) first elaborate on the motivation behind ODIT, (ii) then present a modification for ODIT to prove its asymptotic optimality in the minimax sense under certain conditions, (iii) extensively analyze its computational complexity, (iv) propose an extension of ODIT for the cases where training data is available for some anomaly settings, (v) introduce a unified framework for the proposed ODIT detectors, (vi) and finally provide a simulation study to exemplify the timely and accurate detection by the proposed detectors under a challenging scenario in which univariate methods fail.

The rationale behind using $k$NN distance for anomaly detection is the similarity between the inverse $k$NN distance and likelihood. Specifically, for $f(\vx_i) \ge f(\vx_j), ~\vx_i, \vx_j \in \mathcal{X}$, it is expected that the distance $g_k(\vx_i)$ of $\vx_i$  to its $k$th nearest neighbor in $\cX$ is smaller than that of $\vx_j$. This probability increases with the size of $\cX$, i.e., $\lim_{|\cX|\to\infty} P(g_k(\vx_i) \le g_k(\vx_j)) = 1$.
This in turn provides grounds for using the difference of $k$NN distances in ODIT to approximate the log-likelihood ratio $\ell_t$.

\begin{figure}
\vspace {-0.2in}
\begin{center}
	\begin{subfigure}[h]{0.8\linewidth}
	\includegraphics[width=\linewidth]{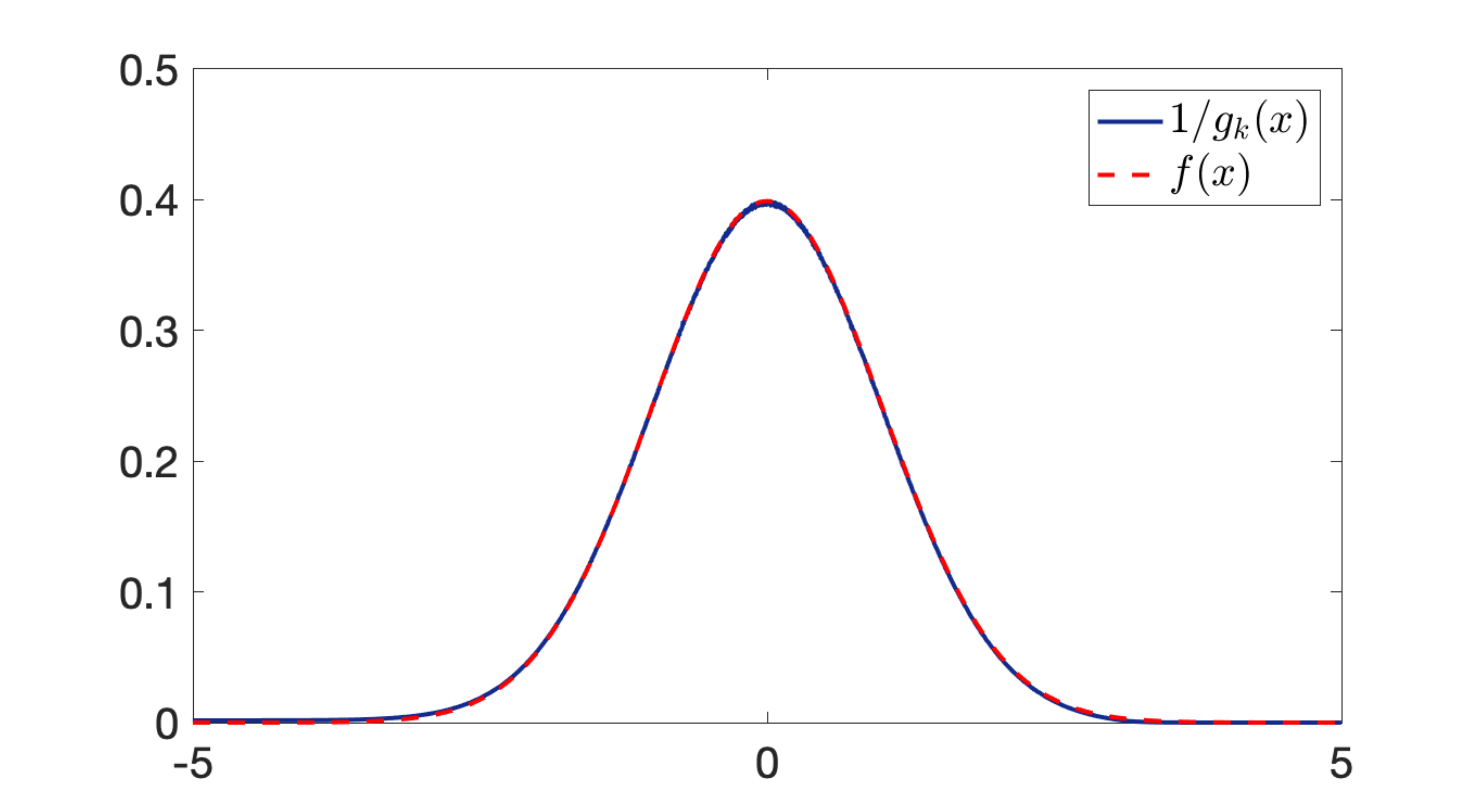}
	\caption{\label{fig:gauss} Standard Gaussian distribution}	
	\end{subfigure}
	\begin{subfigure}[h]{0.8\linewidth}
	\includegraphics[width=\linewidth]{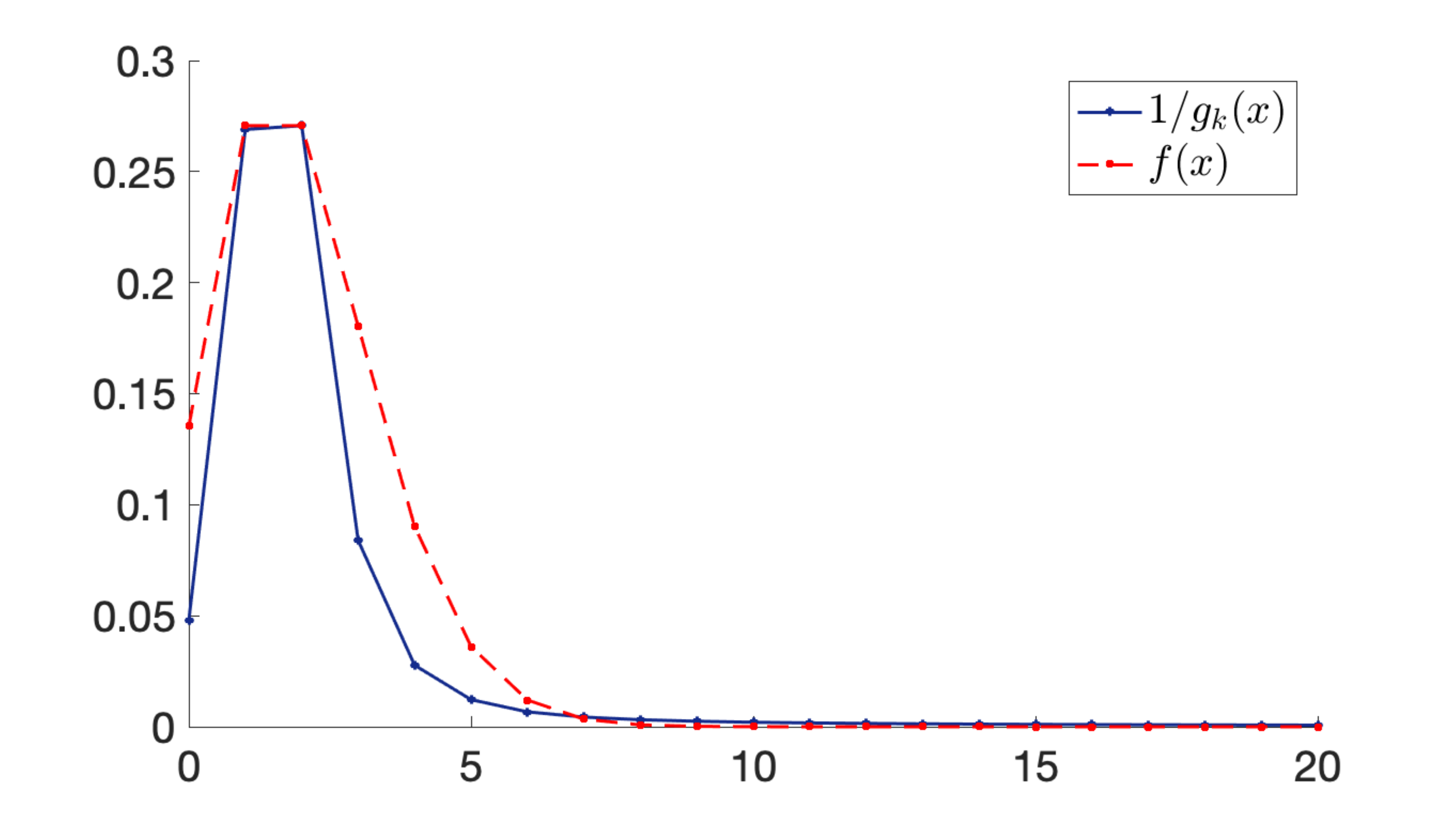}
	\caption{Poisson distribution, $\lambda=2$}
	\end{subfigure}
	\begin{subfigure}[h]{0.8\linewidth}
	\includegraphics[width=\linewidth]{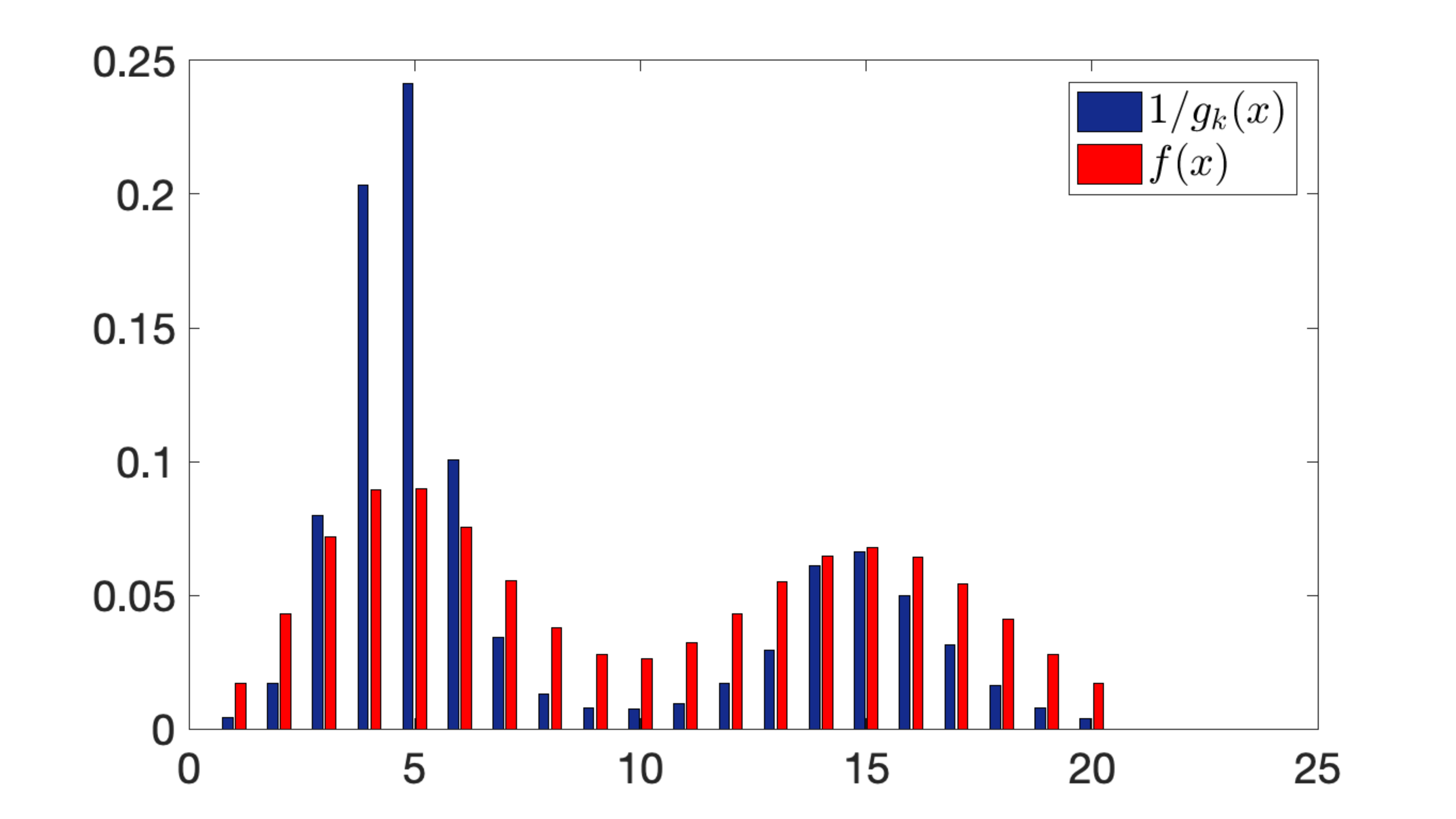}
	\caption{Multinomial distribution}
	\end{subfigure}	
		\begin{subfigure}[h]{0.8\linewidth}
	\includegraphics[width=\linewidth]{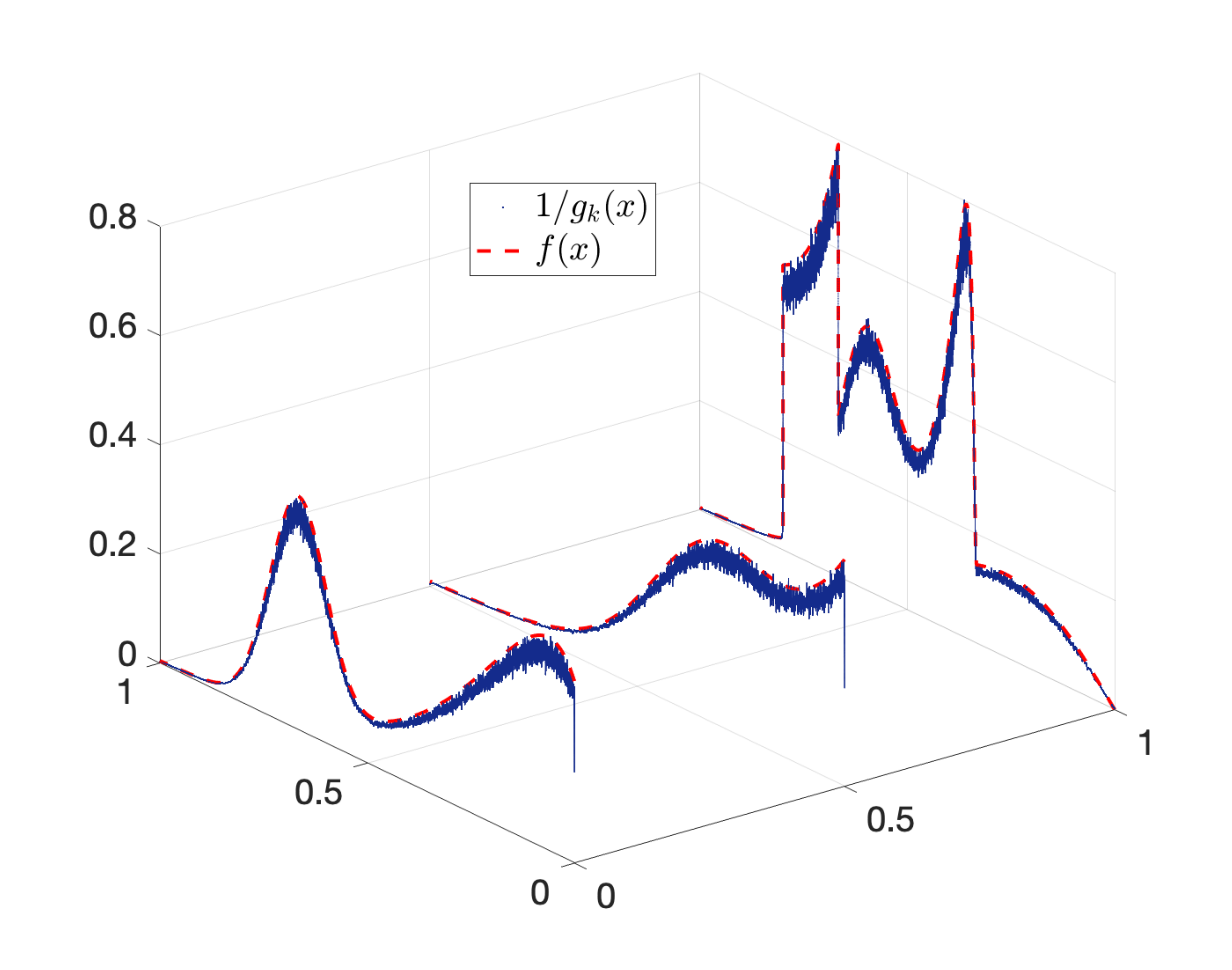}
	\caption{A complex 2d distribution}
	\end{subfigure}		
	\caption{Similarity between inverse $k$NN distance $1/g_k(x)$ and likelihood $f(x)$ for $k=1$.}	
	\label{fig:gauss_knn}
\end{center}
\vspace {-0.35in}
\end{figure}

The similarity between the likelihood of data points and the inverse $k$NN distance is shown in Fig. \ref{fig:gauss_knn} for several distributions. We consider Gaussian, Poisson and multinomial distributions to illustrate the similarity of $1/g_k(x)$ and $f(x)$ for three disparate data types, real-valued numeric, integer-valued numeric and categorical, respectively. The inverse $k$NN distance graphs are scaled down to match the likelihood figure for the purpose of visualization. As shown in Fig. \ref{fig:gauss_knn}(a) with $|\cX|=10^6$, the inverse of $k$NN distance approximates the likelihood very well for the standard Gaussian random variable. Despite some discrepancy for the Poisson and multinomial cases due to the discreteness of these random variables, it may still serve well the purpose of approximating the log-likelihood ratio. For these discrete cases, to avoid zero $k$NN distance we consider much smaller number of data points, $10$ and $50$ for Poisson and multinomial, respectively. Fig. \ref{fig:gauss_knn}(b) and (c) are obtained by averaging over $5\times 10^5$ and $10^4$ trials, respectively. 
In order to show the similarity for a more complex distribution, in Fig. \ref{fig:gauss_knn}(d) we consider a two-dimensional vector of a categorical random variable and a real-valued random variable with arbitrary distribution and $10^4$ data points.

\subsection{Online Discrepancy Test (ODIT)}

\begin{figure}[t]
\vspace{-0.1in}
\begin{center}
	\includegraphics[width=0.5\textwidth]{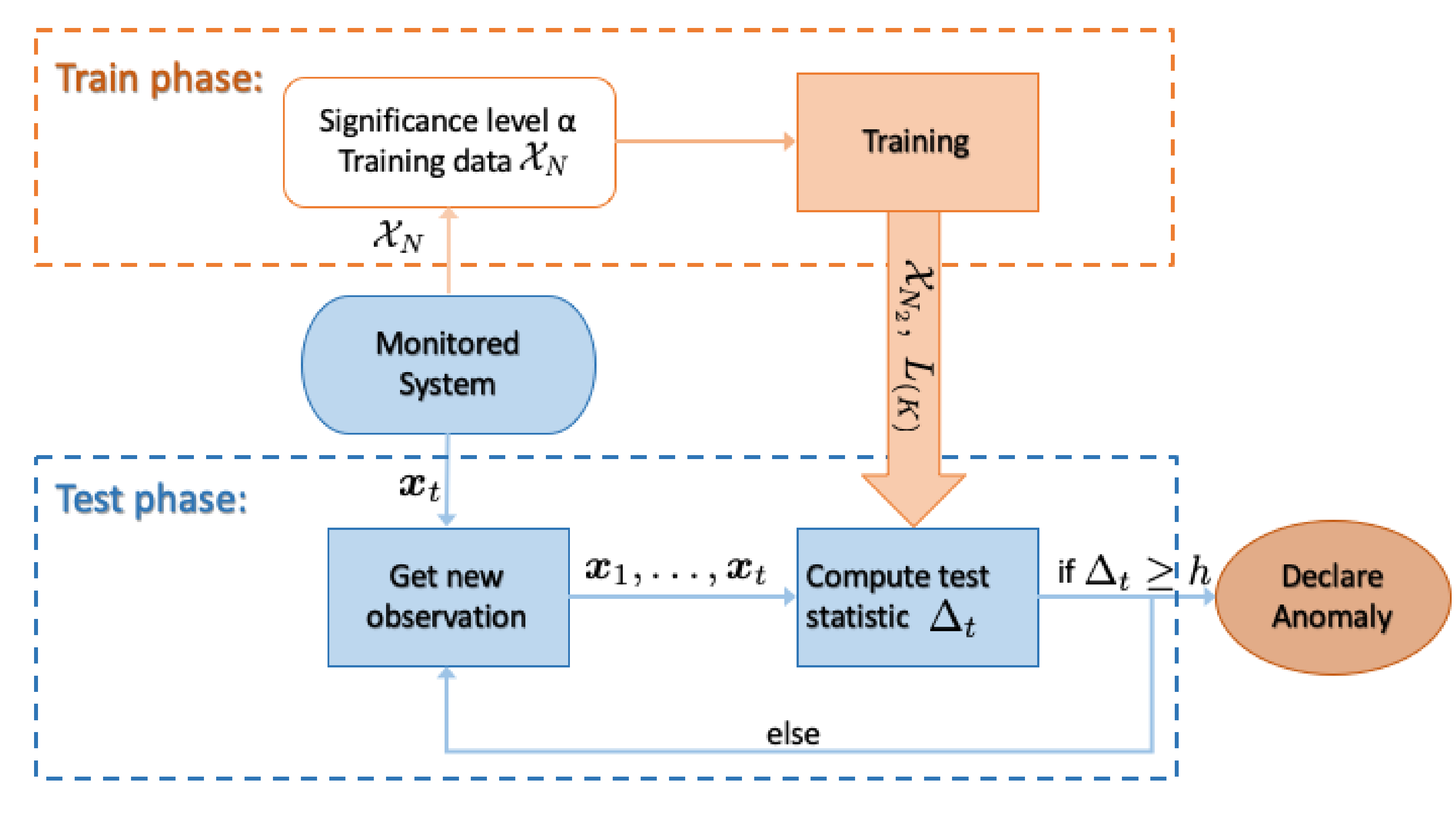}
	\caption{Overview of the ODIT anomaly detection mechanism.}
	\label{fig:overview}
\end{center}
\vspace{-0.2in}
\end{figure}

The overview of the ODIT detector is given in Fig. \ref{fig:overview}. 
In the training phase, assuming a training set $\cX_N$ consisting of $N$ nominal data instances, it firstly partitions $\cX_N$ into two sets $\cX_{N_1}$ and $\cX_{N_2}$, where $N_1+N_2=N$, for computational efficiency as in the bipartite GEM algorithm \cite{GEM-2}. Then, using the $k$NN distances $\{g_k(\vx_m)\}$ between each node $\vx_m \in \cX_{N_1}$ and its $k$ nearest neighbors in $\cX_{N_2}$ ODIT finds an estimate $\hat{\Omega}_\alpha$ for the minimum volume set (MVS) $\Omega_\alpha$ given by
\be
\Omega_\alpha = \argmin_{\cA}{\int_\cA \td \vx}~~\text{s.t.}~~\int_\cA f_0(\vx) \td \vx \geq 1 - \alpha,
\ee
where $\alpha \in (0,1)$ is a significance level, e.g., $0.05$. $\Omega_\alpha$ represents the most compact set of observations under nominal operation while its complement $\overline{\Omega}_\alpha$ corresponds to the tail events (i.e., outliers) under nominal operation at significance level $\alpha$. 
Then, in the test phase, it compares the $k$NN distances $g_k(\vx)$ between a test data instance $\vx$ and its $k$ nearest neighbors in $\cX_2$ with $\hat{\Omega}_\alpha$ to compute a negative/positive anomaly evidence for anomaly $\vx$ and accumulates it over time for reliable detection. Roughly, the greater $g_k(\vx)$ is, the less likely $\vx$ comes from the same distribution $f_0$ as the nominal points. The estimate $\hat{\Omega}_\alpha$ provides a reference to evaluate $g_k(\vx)$ and compute the negative/positive anomaly evidence for $\vx$. 

Specifically, in the training phase, to estimate $\Omega_\alpha$ ODIT ranks the points in $\cX_{N_1}$ in the ascending order $\{\vx_{(1)},\ldots,\vx_{(N_1)}\}$ in terms of the total distance
\begin{equation}
\label{e:length}
	L_m = \sum_{n=k-s+1}^k g_n(\vx_m)^\gamma,
\end{equation}
where $g_n(\vx_m)$ is the Euclidean distance between point $\vx_m\in\cX_{N_1}$ and its $n$th nearest neighbor in $\cX_{N_2}$, $s \in [1, k]$ is a fixed number introduced for convenience, and $\gamma>0$ is the weight. Next, it picks the first $K$ points $\cX_{N_1}^K=\{\vx_{(1)},\ldots,\vx_{(K)}\} \subset \cX_{N_1}$ with the smallest total distances $\{L_{(1)},\ldots,L_{(K)}\}$ to estimate the MVS $\Omega_\alpha$, i.e., $\hat{\Omega}_\alpha=\cX_{N_1}^K$. 
It is known \cite{GEM-2} that $\cX_{N_1}^K$ converges to $\Omega_\alpha$ as 
$$
	\lim_{K,N_1 \to \infty} K/N_1 \to 1-\alpha.
$$
Hence, $K$ is chosen as $K= \lfloor N_1 (1-\alpha) \rfloor$, where $\lfloor \cdot \rfloor$ is the floor operator. 

In the test phase, for each data instance $\vx_t$, ODIT firstly computes the total distance $L_t$ with respect to the second training set $\cX_{N_2}$ as in \eqref{e:length}. Then, it computes the anomaly evidence, which could be either positive or negative, by comparing $L_t$ with the MVS model found in the training phase through the borderline total distance $L_{(K)}$
\be
\label{eq:evidence}
D_t = d (\log L_t - \log L_{(K)}), 
\ee
where $d$ is the number of data dimensions. Finally, it updates a detection statistic $\Delta_t$ which accumulates the anomaly evidence $D_t$ over time, and raises an anomaly alarm the first time $\Delta_t$ crosses a predefined threshold, 
\begin{align} 
\label{eq:delta_update}
\begin{split}
\Delta_t &= \max\{\Delta_{t-1} + D_t, 0\}, \quad \Delta_0 = 0, \\
T &= \min\{t : \Delta_t \geq h\}, 
 \end{split}
\end{align}
which is a CUSUM-like procedure (cf. \eqref{eq:cusum_proc}). The ODIT procedure is summarized in Algorithm \ref{alg:odit1}.

\begin{algorithm}
\caption{The proposed ODIT procedure}
\label{alg:odit1}
\begin{algorithmic}[1]
\State \emph{Input:} $\cX_N,k,s,\alpha,h$
\State \emph{Initialize:} $\Delta \gets 0, \text{t} \gets 1$
\State \emph{Training phase:}
\State Partition $\cX_N$ into two sets $\cX_{N_1}$ and $\cX_{N_2}$
\State For each $\vx_m \in \cX_{N_1}$ compute $L_m$ as in \eqref{e:length}
\State Find $L_{(K)}$ by selecting the $K$th smallest $L_m$
\State \emph{Test phase:}
\While {$\Delta < \textit{h}$}
\State Get new data $\vx_t$ and compute $D_t$ as in \eqref{eq:evidence}
\State $\Delta = \max\{\Delta+D_t,0\}$
\State $t \gets t + 1$
\EndWhile
\State $\text{Declare Anomaly}$
\end{algorithmic}
\end{algorithm}

The computation of the anomaly evidence $D_t$ for each test instance $\vx_t$ has the simpler form $D_t = L_t - L_{(K)}$ in \cite{odit}, where we proposed ODIT the first time. Although this simpler form of $D_t$ and the form proposed in \eqref{eq:evidence} have similar structures, and they perform quite similarly in practice, the new form given in \eqref{eq:evidence} naturally appears while proving the asymptotic optimality of ODIT in the minimax sense, as shown next.

\begin{theorem}
\label{thm:odit}
When the nominal distribution $f_0(\vx_t)$ is finite and 
continuous, and the attack distribution $f_1(\vx_t)$ is a uniform distribution, as the training set grows, the ODIT statistic $D_t$ converges in probability to the log-likelihood ratio,
\begin{equation}
D_t \overset{p}{\to} \log \frac{f_1(\vx_t)}{f_0(\vx_t)}
~~~ \text{as}~~~ N_2 \to \infty,
\end{equation}
i.e., ODIT converges to CUSUM, which is minimax optimum in minimizing expected detection delay while satisfying a false alarm constraint.
\end {theorem}

\begin{IEEEproof}
Consider a hypersphere $\cS_t \in \bR^d$ centered at $\vx_t$ with radius $g_k(\vx_t)$, the $k$NN distance of $\vx_t$ with respect to the training set $\cX_{N_2}$. The maximum likelihood estimate for the probability of a point being inside $\cS_t$ under $f_0$ is given by $k/N_2$. It is known that, as the total number of points grow, this binomial probability estimate converges to the true probability mass in $\cS_t$ in the mean square sense \cite{agresti2018introduction}, i.e., $k/N_2 \overset{L^2}{\to} \int_{\cS_t} f_0(\vx) ~\text{d}\vx$ as $N_2 \to \infty$. Hence, the probability density estimate $\hat{f}_0(\vx_t)=\frac{k/N_2}{V_d g_k(\vx_t)^d}$, where $V_d g_k(\vx_t)^d$ is the volume of $\cS_t$ with the appropriate constant $V_d$, converges to the actual probability density function, $\hat{f}_0(\vx_t) \overset{p}{\to} f_0(\vx_t)$ as $N_2 \to \infty$, since $\cS_t$ shrinks and $g_k(\vx_t) \to 0$. Similarly, considering a hypersphere $\cS_{(K)} \in \bR^d$ around $\vx_{(K)}$ which includes $k$ points with its radius $g_k(\vx_{(K)})$, we see that as $N_2 \to \infty$, $g_k(\vx_{(K)})\to 0$ and $\hat{f}_0(\vx_{(K)})=\frac{k/N_2}{V_d g_k(\vx_{(K)})^d} \overset{p}{\to} f_0(\vx_{(K)})$. Assuming a uniform distribution $f_1(\vx)=f_0(\vx_{(K)}), ~\forall \vx$, we conclude with
$\log \frac{\frac{k/N_2}{V_d g_k(\vx_{(K)})^d}}{\frac{k/N_2}{V_d g_k(\vx_t)^d}} = d \left[ \log g_k(\vx_t) - \log g_k(\vx_{(K)}) \right] \overset{p}{\to} \log \frac{f_1(\vx_t)}{f_0(\vx_t)} ~~ \text{as}~~ N_2\to\infty$,
where $L_t=g_k(\vx_t)$ for $s=\gamma=1$. For $\gamma$ values different than $1$, $D_t$ converges to the log-likelihood ratio scaled by $\gamma$. 
\end{IEEEproof}

Note that ODIT does not train on any anomalous data, i.e., does not use any knowledge of anomaly to be detected. While this generality is an attractive trait as it allows detection of any statistical anomaly, it also inevitably limits the performance for known anomaly types on which detectors can train. We will extend ODIT to this case with available anomaly information in Section \ref{sec:odit2}. In Theorem \ref{thm:odit}, we show that in the lack of knowledge about anomalies, ODIT reasonably assumes an uninformative uniform likelihood for the anomaly case, and achieves asymptotic optimality under this assumption in the CUSUM-sense for certain parameter choices. 

\emph{Remark 1 (Parameter Selection):}
Due to its sequential nature, the parameters of ODIT either directly or indirectly control the fundamental trade-off between minimizing average detection delay and false alarm rate.
Parameters $k$ and $s$ determine how many nearest neighbors to take into account in computing the total distance $L_m$, given by \eqref{e:length}. Smaller $k$ would result in being more sensitive to anomaly, hence supports earlier detection, but at the same time it causes to be more prone to the false alarms due to nominal outliers. Larger $k$ would result in vice versa. $s$ is an auxiliary parameter chosen for further flexibility in this trade-off. $s=1$ considers only the $k$th nearest neighbor while $s=k$ sums all the first $k$ nearest neighbors. Similar to $k$, smaller $s$ makes the algorithm more sensitive to anomaly, but also more prone to nominal outliers. However, the effect of $s$ is secondary to that of $k$. $k$ and $s$ should be chosen together to strike a balance between sensitivity to anomalies and robustness to nominal outliers. $0 < \gamma < d$ is the weight which determines the emphasis on the difference between distances. Large distance values are emphasized by large $\gamma$ values and suppressed by small $\gamma$ values.
The alarm threshold $h$ in \eqref{eq:delta_update} directly controls the trade-off between minimizing detection delay and false alarm rate. Decreasing $h$ will yield smaller detection delays, i.e., earlier detection, but also more frequent false alarms. It is typically selected to satisfy a false alarm constraint. 
The significance level $\alpha$ is at a secondary role supporting $h$. For fixed $h$, larger $\alpha$ would result in a smaller estimated MVS $\hat{\Omega}_\alpha$, which in turn results in smaller detection delays, but also more frequent false alarms since more nominal data points will lie outside the selected MVS. Note that $h$ is the final decision threshold, whereas $\alpha$ is more of an intermediate parameter. Hence, one can always set $\alpha$ to a reasonable significance value, such as $0.05$, and then adjust $h$ accordingly to satisfy a desired false alarm rate. 
Regarding the sizes of training sets $N_2$ plays a more important role than $N_1$, as shown in Theorem \ref{thm:odit}. Specifically, $N_2$ determines the accuracy of likelihood estimates by the $k$NN distances, whereas $N_1$ determines how well the significance level $\alpha$ is satisfied, which is an intermediate parameter as discussed before. Hence, typically $N_2$ should be chosen larger than $N_1$, where $N_1+N_2=N$. 
It should be noted that the ODIT procedure, given by Algorithm \ref{alg:odit1}, can also work without partitioning the training set. Partitioning is proposed for computational efficiency when dealing with large high-dimensional datasets. However, it does not decrease the order of magnitude in computational complexity (see Section \ref{sec:comp_complex}) since even without partitioning the online testing procedure already scales linearly with the number of training instances, as opposed to the bipartite GEM algorithm \cite{GEM-2} which decreases the complexity to linear from exponential using partitioning. 
As a result, Algorithm \ref{alg:odit1} can be used without partitioning the training set, especially for small datasets. 

\emph{Remark 2 (Graph Interpretation):}
The $K$ points in MVS estimate $\cX_{N_1}^K$ and their $k$ nearest neighbors in $\cX_{N_2}$ form an Euclidean $k$NN graph $\cG=(\overline{\cX}_{N_1}^K,\cE)$, where $\overline{\cX}_{N_1}^K$ is the set of vertices and $\cE$ is the set of edges connecting $\cX_{N_1}^K$ to the neighbors in $\cX_{N_2}$. 
The constructed graph $\cG$ minimizes the total edge length $\sum_{m=1}^K L_m$ among all possible $K$-point $k$NN graphs between $\cX_{N_1}$ and $\cX_{N_2}$. The computation of anomaly evidence $D_t$ in \eqref{eq:evidence} can then be interpreted as the increase/decrease in the log of total edge length if the $K$-$k$NN graph were to include the test point $\vx_t$.

\emph{Remark 3 (Comparisons):}
ODIT learns $\hat{\Omega}_\alpha$ using $k$NN distances similarly to the outlier detection method called Geometric Entropy Minimization (GEM) \cite{GEM,GEM-2}. However, in the test phase, unlike GEM, which declares anomaly even when a single test point falls outside the MVS, ODIT sequentially updates a test statistic $\Delta_t$ using the closeness/remoteness of the test point to the MVS, and declares anomaly only when $\Delta_t$ is large enough, i.e., there is enough anomaly evidence with respect to a false alarm constraint. Doing so ODIT is able to timely and accurately detect persistent anomalies, as shown theoretically in Theorem \ref{thm:odit} and through numerical results in Section \ref{sec:sim-correlation} and Section \ref{sec:ddos}. Whereas, one-shot outlier detectors like GEM are prone to high false alarm rates due to the limitation of significance tests \cite{pval1,pval2}. 
The sequential detection structure of ODIT resembles that of CUSUM albeit with fundamental differences. Actually, the test statistic of ODIT implements a discrepancy function motivated by the discrepancy theory \cite{dtheory} and discrepancy norm \cite{dnorm}, hence the name \emph{Online Discrepancy Test (ODIT)}. The nonparametric nature of ODIT does not require any knowledge of the nominal and anomaly probability distributions, as opposed to CUSUM. Moreover, the practical relaxations of CUSUM, such as G-CUSUM and independent CUSUM \cite{MEI}, cannot be applied to challenging scenarios such as high-dimensional systems which require multivariate anomaly detection with little or no knowledge of anomaly types. On the other hand, ODIT scales well to high-dimensional systems for multivariate detection, as discussed next.

\subsection{Computational Complexity}
\label{sec:comp_complex}

Next, we analyze the computational complexity of our proposed method. Training phase of ODIT requires the $k$NN distances between each pair of the data points in the two training sets. Therefore, the time complexity of training phase is $O(N_1 N_2 d)$, where $d$ is the data dimensionality. The space complexity of training is $O(N_2 d)$ since $N_2$ points are stored for testing. Note that training is performed once offline, thus the complexity of online testing is usually critical for scalability. In the test phase, computing the $k$NN distance of a test point among all points in the second training set takes $O(N_2 d)$ time. The space complexity of testing is not significant as the test statistic is updated recursively. Consequently, the proposed ODIT algorithm linearly scales with the data dimensionality $d$ both in training and testing. In the online testing phase, it also scales linearly with the number of training points. For high-dimensional systems with abundance of training data, the online testing time could be the bottleneck in implementing ODIT. 

\emph{$k$NN Approximation:} Computing the nearest neighbors of a query point is the most computationally expensive part of the algorithm as the distance to every other point in the second training data needs to be computed to select the $k$ smallest ones. As the dimensionality increases and the training size grows, the algorithm becomes less efficient in terms of the running time.  To this end, we propose to approximate the $k$NN distance rather than computing its exact value. It is natural to expect that ODIT's performance will drop due to the inaccuracy induced by the approximated $k$NN distances compared to that based on the exact $k$NN distances. However, depending on the system specifications, e.g., how frequently the data arrives and how critical timely detection is, the reduction in running time through $k$NN approximation may compensate for the performance loss, as we next analyze through an experiment. \cite{Muja} proposes a $k$NN distance approximation algorithm that scales well to high-dimensional data. This algorithm performs hierarchical clustering by constructing a k-means tree, and approximates the $k$NN distance by performing a priority search in the k-means tree, i.e., by searching for the $k$ nearest neighbors only among a limited number of data points. 
The computation complexity of constructing the tree is $O(N_2 d C I_{max} \frac{\log N_2}{\log C})$, where $I_{max}$ is the maximum number of iterations in k-means clustering, $C$ is the number of clusters (a.k.a. branching factor), and $\frac{\log N_2}{\log C}$ is the average height of the tree. 
Using the priority search k-means tree algorithm, the computational complexity of $k$NN search reduces to $O(B d \frac{\log N_2}{\log C})$, where $B \ll N_2$ is the maximum number of data points to examine. Hence, the training complexity reduces to $O((N_1 B+N_2 C I_{max})\frac{\log N_2}{\log C} d)$ from $O(N_1 N_2 d)$. Note that $B \ll N_2$ and the number of iterations required for convergence is small \cite{Muja}. More importantly, in online testing, the computational complexity per instance decreases to $O(B \frac{\log N_2}{\log C} d)$ from $O(N_2 d)$.

\emph{Experiment:} We experimented with this approximation in our algorithm. The experiment is done in Matlab on an Intel 3.60 GHz processor with 16 GB RAM. In the experiment, the dimensionality of data is $d=50$, the training data size is $N=5\times 10^5$, partitioned into $N_1 = 0.38 N$ and $N_2 = 0.62 N$ \footnote{The same partitioning ratio is used in the experiments throughout the paper.}, and the anomaly is defined as a shift in the mean of Gaussian observations by 3 standard deviation in $10\%$ of the dimensions. We set the branching factor for building the priority search k-means tree as $C=100$, and the maximum number of points to examine during search for the $k$ nearest neighbors as $B=1000$. The average computation time for both ODITs based on the exact and the approximate $k$NN distance is summarized in the Table \ref{tab:table2}, which presents the time spent for the computation of \eqref{eq:evidence} and \eqref{eq:delta_update} per observation. It is seen that the approximation method drops the average running time per observation to about $1/14$ of that of the exact method.

\begin{figure}[t]
\begin{center}	
	\includegraphics[width=\linewidth]{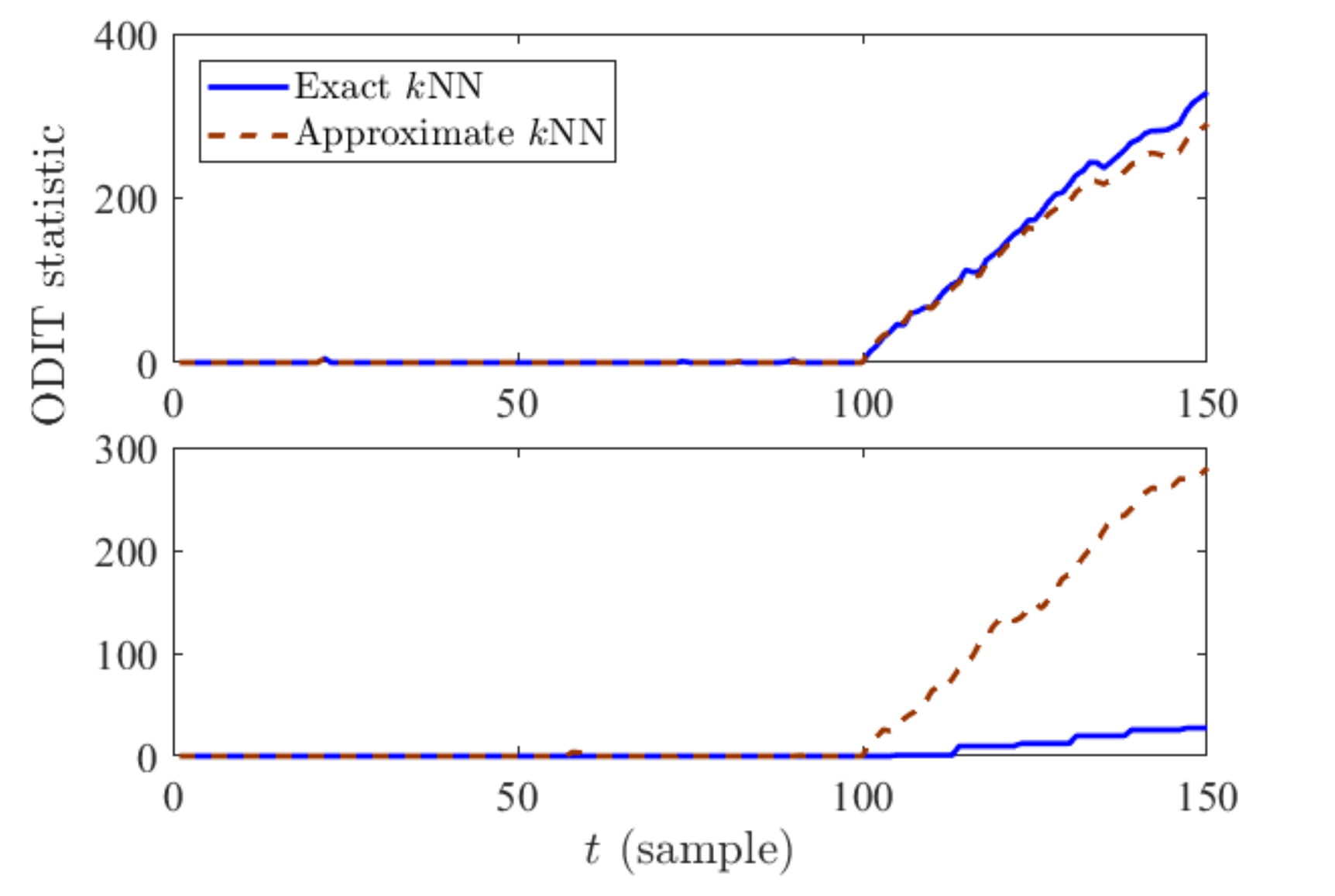}
	\caption{\label{fig:complexity} ODIT statistics based on exact and approximate $k$NN distances when $T_{\text{sampling}} = 1$ sec. (top) and $T_{\text{sampling}} = 0.01$ sec. (bottom).}	
\end{center}
\vspace{-0.2in}
\end{figure}

\begin{table}[h!]
  \begin{center}
    \caption{Average computation overhead of original ODIT and efficient ODIT per sample}
    \label{tab:table2}
    \begin{tabular}{|c|c|}
    \hline
     \multicolumn{2}{|c|}{\bf{Average execution time (sec.)}} \\ \cline{1-2}
      \textbf{Exact $k$NN} & \textbf{Approximate $k$NN}  \\
      \hline
       0.0750 & 0.0054\\
       \hline
      \end{tabular}
  \end{center}
  \vspace{-0.2in}
\end{table}

\begin{figure}[t]
\begin{center}	
	\includegraphics[width=\linewidth]{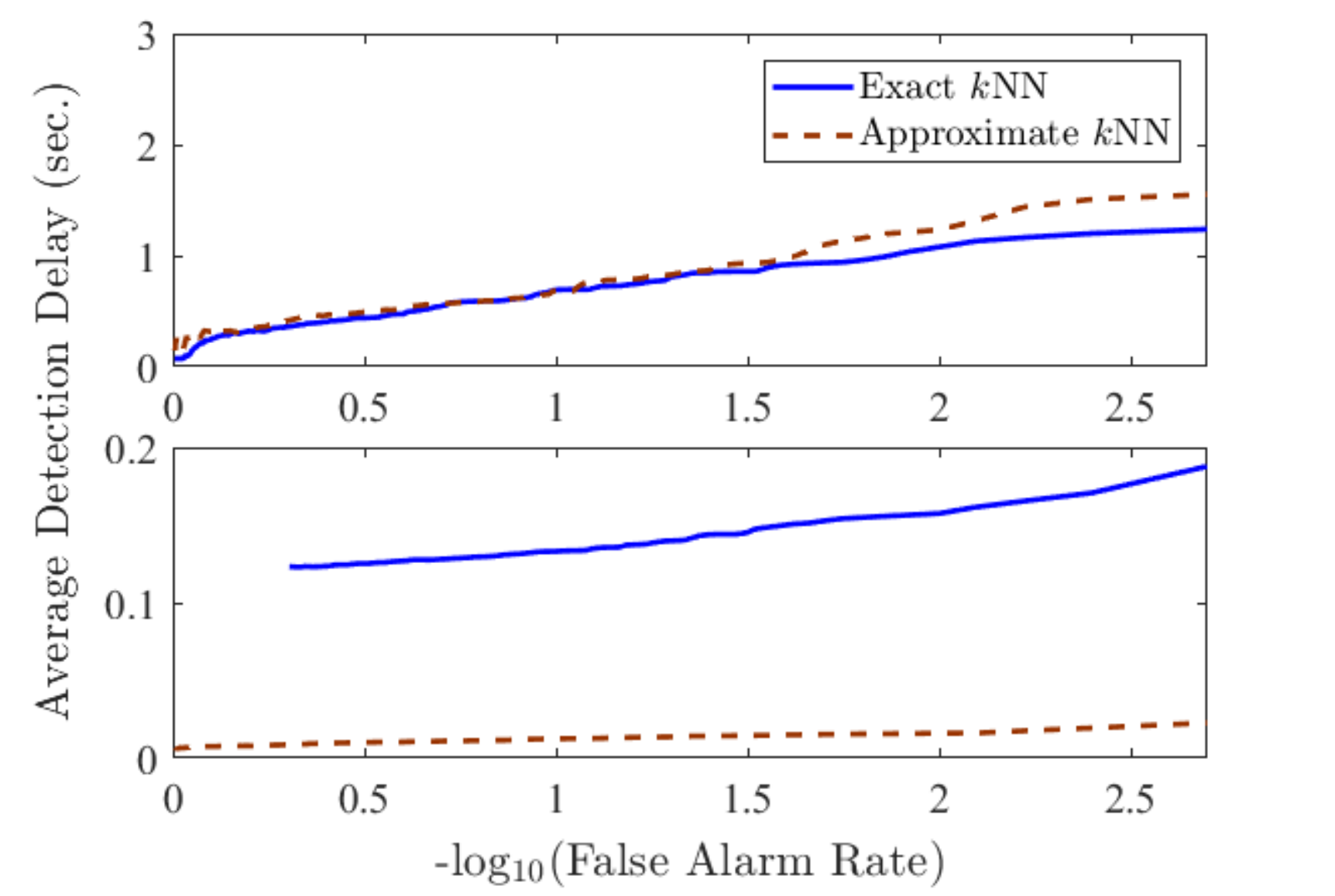}
	\caption{\label{fig:sampling_freq_compr} Comparison between performance of ODIT based on exact and approximate $k$NN distances in terms of seconds for $T_{\text{sampling}} = 1$ sec. (top) and $T_{\text{sampling}} = 0.01$ sec. (bottom).}
	\end{center}
\vspace{-0.2in}
\end{figure}

To compare the original and efficient ODITs in systems with different specifications, in terms of the frequency of data arrival, we considered the following two scenarios: (i) data arrives every $1$ sec., and (ii) data arrives every $0.01$ sec. Figs. \ref{fig:complexity} and \ref{fig:sampling_freq_compr}  compare the decision statistics and average performance of ODIT based on exact and approximate $k$NN in the two scenarios. 
Considering the extra samples needed for detection after the attack onset, as well as the computation time overhead for the last sample before detection, the actual detection delay in time unit is given by $sample~delay \times sampling~period+computational~overhead$. Depending on the sampling period, either exact $k$NN or approximate $k$NN could be more advantageous. For a sampling period that is smaller than the computation overhead, exact $k$NN computations are usually not feasible, causing the original ODIT to miss multiple samples while performing the test for a data instance, as can be seen in the staircase statistic in solid blue in the bottom figure of Fig. \ref{fig:complexity}. Therefore, in such a case, approximate $k$NN computations are preferred over the exact $k$NN computations in terms of the actual detection delay (see the bottom figure in Fig. \ref{fig:sampling_freq_compr}). Whereas for a sufficiently large sampling period, the delay is mainly due to the extra samples, thus exact $k$NN computations yield better results this case, as shown in the top figure in Fig. \ref{fig:sampling_freq_compr}.

\ignore{
\begin{table}[h!]
  \begin{center}
    \caption{comparison of ODIT performance with exact knn versus fast knn in terms of average delay, and execution time}
    \label{tab:table1}
    \begin{tabular}{c||c c||c c}
    & \multicolumn{2}{c||}{\bf{Exact knn}} & \multicolumn{2}{c}{\bf{Fast knn}}\\ \cline{2-5}

      \textbf{Dimensionality} & \textbf{ time(s)} & \textbf{delay(sample)} & \textbf{ time(s)} & \textbf{delay(sample)} \\
      \hline

      50 & 0.1455 & 0.6263 & 0.0085 & 1.6212\\
      80 & 0.0965 & 0.2043 & 0.0081 & 1.8158\\ 
      100 & 0.1301 & 0.0947 & 0.0098 & 1.7141\\ 
    \end{tabular}
  \end{center}
\end{table}
}

\emph{Summary of ODIT:} Here we highlight the prominent features of the proposed ODIT anomaly detector:
\begin{itemize}
	\item The \emph{sequential} nature of ODIT makes it suitable for \emph{real-time} systems, and especially for systems in which \emph{quick and accurate} detection is critical. Additionally, as the nominal training set grows, it asymptotically achieves the minimax optimality in terms of quick and accurate detection when anomaly is from uniform distribution. 
	\item It is capable of performing \emph{multivariate} detection in \emph{high-dimensional} systems, as illustrated in Section \ref{sec:sim-correlation}, thanks to its \emph{nonparametric} and \emph{scalable} nature. 
	\item ODIT can detect \emph{unknown anomaly} types since it does not depend on any assumption about anomalies. Moreover, it is suitable for \emph{online learning} such that its detection performance can be improved over time for previously encountered anomaly types (see Section \ref{sec:ODIT-uni}). 

\end{itemize}

\subsection{An Extension: ODIT-2}
\label{sec:odit2}

In this section we consider the case of having an additional anomaly training dataset along with the previously discussed nominal dataset. Next, we  extend the ODIT method to take advantage of the anomaly dataset in order to improve its performance. With the inclusion of an anomaly training set, the ODIT-2 procedure is akin to the classification methods based on $k$NN distance \cite{kclass1,kclass2}. However, these methods are not sequential.
Consider an anomaly training set $\cX'_M =\{\vx'_1,\vx'_2,...,\vx'_M\}$ in addition to the nominal set $\mathcal{X}_N = \{\vx_1,\vx_2,...,\vx_N\}$.  
In this case, the anomaly evidence for each instance can be computed by comparing the total distance $L_t$ with respect to the nominal dataset with the total distance $L'_t$ with respect to the anomalous dataset. Thus, there is no need to learn the borderline total distance $L_{(K)}$ in training to be used as a baseline for $L_t$ in testing (cf. \ref{eq:evidence}). That is, no training is needed for ODIT-2. 
However, before testing, a pre-processing might be required to remove the data points that are similar to the nominal train set. The reason for cleaning the anomaly dataset rather than the nominal dataset is that usually anomaly dataset is obtained by collecting observations from a known anomalous event which may typically include nominal observations too. For instance, in a network intrusion detection system (IDS), after occurrence of an attack, several observations could still be of nominal nature. The cleaning step is done by finding and removing the data points of anomaly training set which lie in the estimated MVS of the nominal training set, 
\be
\mathcal{X}^{\text{clean}}_M = \mathcal{X}'_M \setminus \{ \vx'_m \in \cX'_M:  L_{\vx'_m} \leq L_{(K)}\},
\ee
where $ L_{\vx'_m}$ is the total distance of $\vx'_m$ with respect to the nominal points in $\cX_{N_2}$. Hence, the training procedure of ODIT, which finds $L_{(K)}$, can be used for preprocessing the anomalous train data. 

While testing for each test data instance $\vx_t$, the anomaly evidence is calculated by
\be 
\label{eq:odit_D}
D_t = d (\log L_t - \log L'_t) + \log (N/M),
\ee
where $L_t$ and $L'_t$ are the total distances of $\vx_t$ computed using \eqref{e:length} with respect to the points in $\cX_{N_2}$ and $\mathcal{X}^{\text{clean}}_{M_2}$, respectively; and $N$ and $M$ are the number of points in the nominal and (cleaned) anomalous training sets.
The statistic update and decision rule of ODIT-2 are the same as in ODIT, given by \eqref{eq:delta_update}. In the ODIT-2 procedure, different than Algorithm \ref{alg:odit1}, 
\eqref{eq:odit_D} is used in line 9 to compute the anomaly evidence $D_t$.

In practice, there is a typical imbalance between the sizes of nominal and anomaly training sets due to the inherent difficulty of obtaining anomaly samples. Since the total $k$NN distances in a dense nominal set $\cX_N$ are expected to be smaller than those in a sparse anomaly dataset, for an anomalous data point, $L_t$ can be smaller than $L'_t$, resulting in a negative anomaly evidence, which can lead to poor detection. In order to deal with the imbalance of datasets, 
the term $\log (N/M)$ in \eqref{eq:odit_D} acts as a correction factor. Specifically, for $N>M$, $\log (N/M)>0$ compensates for $L_t$ being unfairly small compared to $L'_t$. This correction factor naturally appears in the asymptotic optimality proof, as shown next.


\begin{corollary}
When the nominal distribution $f_0(\vx_t)$ and anomalous distribution $f_1(\vx_t)$ are finite and continuous, as the training sets grow, the ODIT-2 statistic $D_t$, given by \eqref{eq:odit_D}, converges in probability to the log-likelihood ratio,
\begin{equation}
D_t \overset{p}{\to} \log \frac{f_1(\vx_t)}{f_0(\vx_t)}
~~~ \text{as}~~~ M,N \to \infty,
\end{equation}
i.e., ODIT-2 converges to CUSUM, which is minimax optimum in minimizing expected detection delay while satisfying a false alarm constraint.
\end{corollary}

\begin{IEEEproof}
From the proof of Theorem \ref{thm:odit}, we know that $\frac{k/N}{V_d g_k(\vx_t)^d} \overset{p}{\to} f_0(\vx_t)$ as $N \to \infty$. Similarly, we can show that $\frac{k/M}{V_d g'_k(\vx_t)^d} \overset{p}{\to} f_1(\vx_t)$ as $M \to \infty$, where $g'_k(\vx_t)$ is the $k$NN distance of $\vx_t$ in the anomalous training set $\cX'_M$. Hence, we conclude with
$\log \frac{\frac{k/M}{V_d g'_k(\vx_t)^d}}{\frac{k/N}{V_d g_k(\vx_t)^d}} = d \left[ \log g_k(\vx_t) - \log g'_k(\vx_t) \right] + \log(N/M) \overset{p}{\to} \log \frac{f_1(\vx_t)}{f_0(\vx_t)} ~~ \text{as}~~ M,N \to\infty$,
where $L_t=g_k(\vx_t)$ and $L'_t=g'_k(\vx_t)$ for $s=\gamma=1$. 
\end{IEEEproof}

\subsection{Unified Framework for Online Learning}
\label{sec:ODIT-uni}

Availability of labeled training data is a major limiting factor for improving the performance of anomaly detection techniques. In several applications, obtaining a comprehensive and accurate labeled training dataset for the anomaly class is very difficult \cite{SURVEY}. In contrast, in most applications typically sufficient amount of comprehensive nominal training data is available. Semi-supervised techniques including ODIT, constitute a popular class of anomaly detection methods that require labeled training data only for the nominal class. These techniques try to build a model of nominal operation/behavior. Hence, anomaly detection is performed by detecting data which significantly deviates from the constructed nominal model. Supervised techniques on the other hand, assume availability of both nominal and anomalous datasets, and build models for classifying unseen data into nominal vs. anomaly classes. ODIT-2, as an example supervised technique, outperforms the semi-supervised ODIT technique for the known anomaly types, as shown in Section \ref{sec:sim-correlation} and Section \ref{sec:ddos}. However, ODIT-2, and in general supervised anomaly detectors, fall short of detecting unknown anomaly types while ODIT, and in general semi-supervised anomaly detectors, can easily handle new anomaly patterns as they do not depend on assumptions about the anomalies. 

Combining the strengths of ODIT and ODIT-2, we propose an online learning scheme called ODIT-uni which is capable of detecting new anomaly types and and at the same time improving its performance for detecting the previously seen anomaly types. Particularly, in the unified ODIT method, both ODIT and ODIT-2 run in parallel to detect anomalies, and the anomalous data instances first detected by ODIT are included in the anomalous training set of ODIT-2 in order to empower the detection of similar anomaly types. Since the ODIT-2 procedure involves all the necessary elements for ODIT, there is no further computation overhead induced by the unified approach. Keeping track of the cumulative decision statistics of ODIT and ODIT-2 the unified ODIT scheme, ODIT-uni, stops the first time either ODIT or ODIT-2 stops:
\begin{align}
\label{eq:odit_Duni}
\Delta^{(1)}_t &= \max\{\Delta^{(1)}_t + D^{(1)}_t, 0\}, ~ \Delta^{(2)}_t = \max\{\Delta^{(2)}_t + D^{(2)}_t , 0 \} \nonumber\\
T &= \min\{t : \Delta^{(1)}_t \geq h_1 ~\text{or}~ \Delta^{(2)}_t \geq h_2\}, 
\end{align}
where $D^{(1)}_t$ and $D^{(2)}_t$ are the anomaly evidences given by \eqref{eq:evidence} and \eqref{eq:odit_D}, respectively, and $h_1$ and $h_2$ are the decision thresholds for ODIT and ODIT-2. For known anomaly patterns on which ODIT-2 is trained, it is expected that $\Delta^{(2)}_t \ge h_2$ happens earlier, whereas $\Delta^{(1)}_t \ge h_1$ is supposed to detect new anomaly types. If the alarm is raised by ODIT, then the anomaly onset time is estimated as the last time instance the ODIT statistic was zero, i.e., $\hat{\tau} = \max\{ t<T: \Delta^{(1)}_t=0\}$, and the data instances $\{\vx_{\hat{\tau}+1},\ldots,\vx_T \}$ between $\hat{\tau}$ and $T$ are added to the ODIT-2 anomaly training set.
For reliable enhancement of the ODIT-2 anomaly training set with the newly detected instances, the ODIT threshold $h_1$ needs to be selected sufficiently high to prevent false alarms by ODIT, and thus false inclusions into the ODIT-2 training set. Obviously, large $h_1$ will increase the detection delays for previously unseen anomaly types, however, avoiding false training instances is a more crucial objective.

\subsection{Example: Detecting Change in the Covariance of a High-Dimensional System}
\label{sec:sim-correlation}

The nonparametric nature of the proposed ODIT detectors makes them suitable for multivariate detection in high-dimensional and heterogeneous systems. Through an experiment we next show the advantage of ODIT and ODIT-2 over the parametric G-CUSUM detector in a challenging setting where anomaly is manifested as a change in the correlation between the individual data streams. This type of anomaly is well exemplified by the MadIoT attacks, recently introduced in \cite{soltan2018blackiot}, in which high wattage IoT devices, such as air conditioners and water heaters, are synchronously turned on/off to cause instability, and as a result blackout in the power grid. 

\begin{figure}[t]
\begin{center}
	\
	\includegraphics[width=\linewidth]{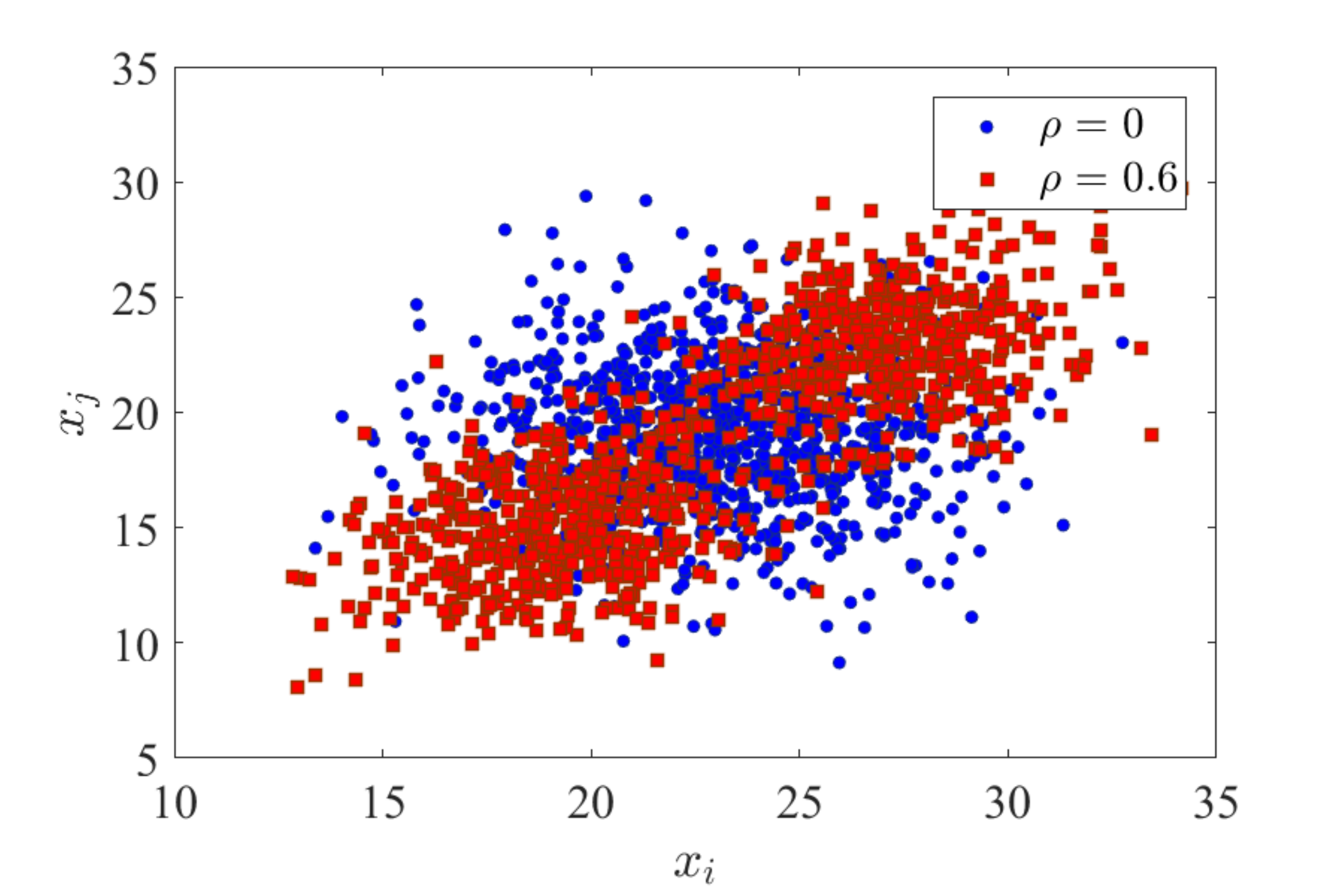}

	\caption{Change in the correlation.\label{fig:corr-data}}
	
\end{center}

\vspace{-0.2in}
\end{figure}

Since it is not tractable to estimate the joint distribution of high-dimensional observations, especially the set of anomalous dimensions in the anomaly case, we implement the G-CUSUM detector proposed in \cite{MEI} which performs univariate analysis by assuming that the data streams are independent. As expected this univariate approach fails to detect the change in the covariance of observations. In the experiment, we simulate a $100$-dimensional system that generates data following a multivariate Gaussian distribution with $\mu = 20$ and $\sigma = 10$ for the individual data streams, which initially have no correlation. At time $t = 100$, the covariance matrix of the observations is changed by randomly adding $\rho=0.6$ correlation between $50\%$ of the data streams without any change in the mean and variance (i.e., diagonal terms in the covariance matrix). Fig. \ref{fig:corr-data} demonstrates the change in the distribution of two data dimensions. For better visualization, some of the anomaly instances that overlap with the nominal instances are not shown.
We used $N=2\times 10^4$ nominal training instances, and $M=2\times 10^4$ anomalous training instances, which decreased to $M=4836$ after cleaning, for a scenario in which $50\%$ of the data dimensions become correlated with $\rho=0.6$.

\begin{figure}[t]
\begin{center}
	\
	\includegraphics[width=\linewidth]{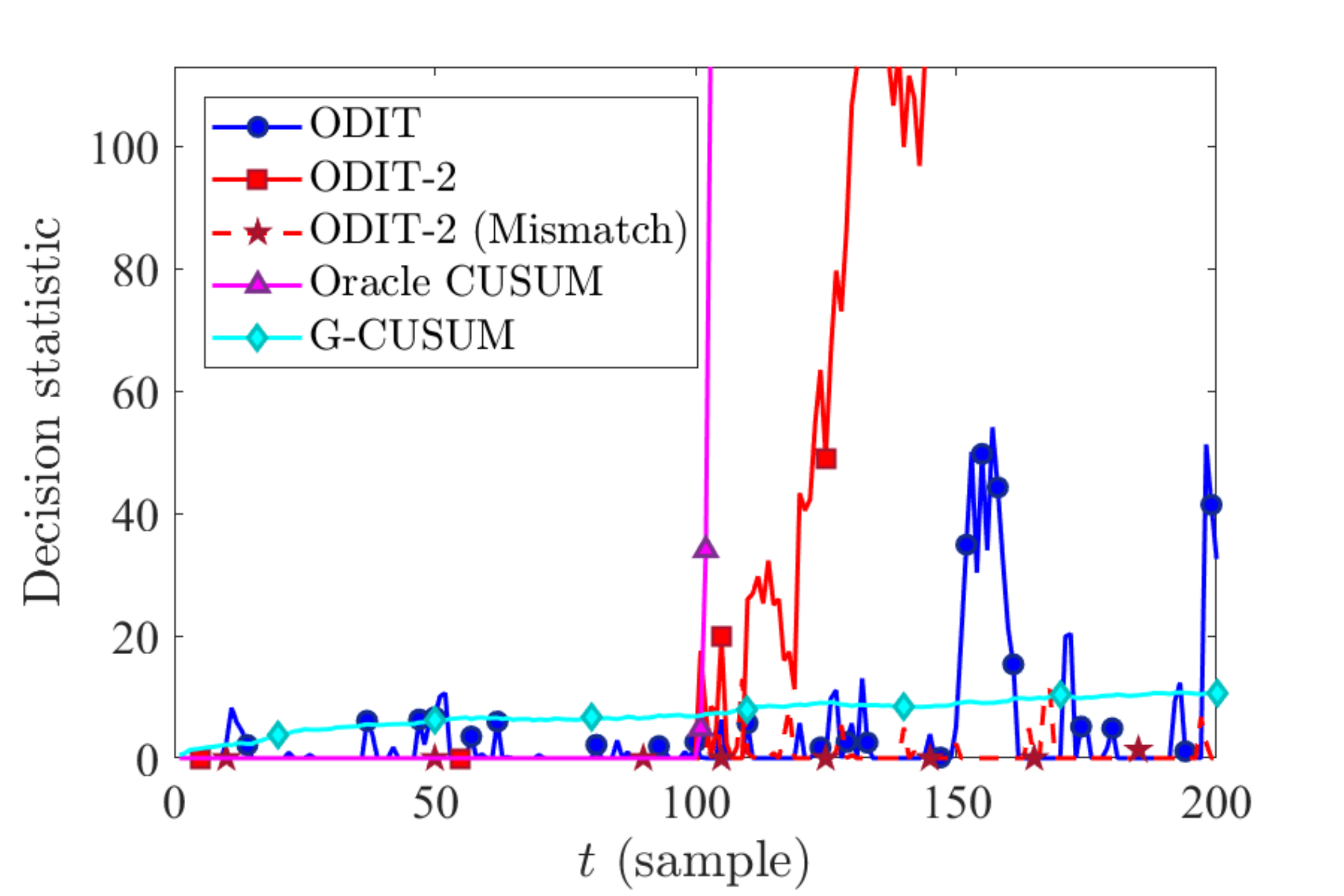}

	\caption{Decision statistics of ODITs vs. G-CUSUM and Oracle CUSUM in the correlation monitoring example.\label{fig:corr-stats}}
	
\end{center}

\vspace{-0.15in}
\end{figure}

\begin{figure}[t]
\begin{center}
	\
	\includegraphics[width=\linewidth]{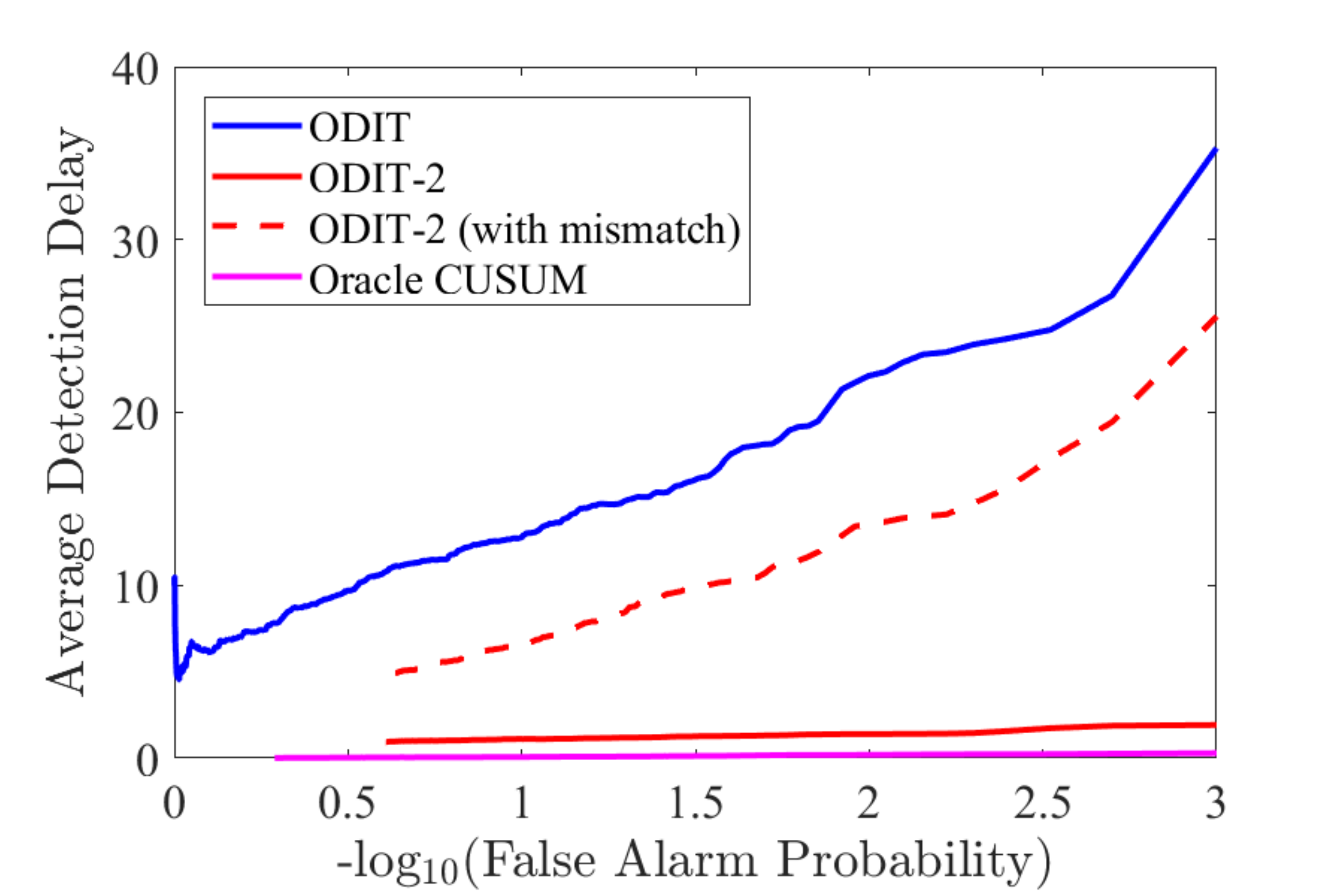}

	\caption{\label{fig:corr-perf} Performance comparison for ODITs in the correlation monitoring example.}
	
\end{center}

\vspace{-0.2in}
\end{figure}

In the experiment we compare the performance of ODIT algorithms with G-CUSUM and Oracle CUSUM, which exactly knows the nominal and anomalous probability distributions. This is a challenging problem due to the fact that the mean and variance of individual data-streams does not change. In particular, some data instances after the anomaly onset are still very similar to the nominal instances. To cope with the similarity of the anomaly instances to the nominal ones, the parameters of ODIT algorithms are set to be $k=s=\gamma=1$, $\alpha_1 = 0.2$, $\alpha_2 = 0.005$, and the cleaning step is performed on the anomaly train set for ODIT-2. 
As depicted by its ever-increasing decision statistic in Fig. \ref{fig:corr-stats}, G-CUSUM fails to detect anomalies since it is not able to monitor correlations. Whereas, the ODIT algorithms successfully detect the change in the covariance structure of observations by performing multivariate analysis, as shown in Fig. \ref{fig:corr-perf}. Since ODIT does not use any anomaly data in training, it detects the anomaly with larger delays compared to ODIT-2. This example provides a scenario where the availability of a set of previously encountered anomalous instances greatly helps ODIT-2 to perform significantly better. ODIT-2 achieves a close performance to the impractical Oracle CUSUM algorithm in the ideal case in which the anomalous dimensions in the test matches the ones in the training. To demonstrate that ODIT-2 is still able to operate reasonably well under non-ideal conditions, we tested it for the case where there is a mismatch between the test and train data in terms of the set of data-streams getting correlated. In this case, 27 out of the 50 dimensions getting correlated are not seen in the anomaly train data. Fig. \ref{fig:corr-perf} shows that despite the mismatch, ODIT-2 still performs better than ODIT.

\section{Anomaly Localization Using ODIT}
\label{sec:localization}

In this section, we propose a localization strategy to identify the data dimensions in which the detected anomaly occurs so that necessary steps can be taken to mitigate the anomaly. Specifically, after an anomaly is detected in ODIT, our objective is to identify the dimensions that caused the detection statistic $\Delta_t$ to increase considerably and ultimately resulted in the detection. Our approach to perform this task is by examining the contribution of each dimension individually to the decision statistics. In the case of detection by ODIT, an increase in the total distance $L_t$, given by \eqref{e:length}, leads to an increase in the anomaly evidence $D_t$, given by \eqref{eq:evidence}, finally leading to an increase in the detection statistic $\Delta_t$, given by \eqref{eq:delta_update}, and consequently the anomaly alarm. Let us assume $\vx_t$ is the test data instance, and $\{\vy_1,\ldots,\vy_k\}$ are its $k$ nearest neighbors in the train set. The total $k$NN distance $L_t = \sum_{n=k-s+1}^k \| \vx_t - \vy_n \|^{\gamma}$, for $\gamma=2$, can be written in terms of the $d$ data dimensions as
\be
\label{eq:L_contr}
L_t = \sum_{i = 1}^d \delta_{t}^i , ~ \text{where}~
\delta_{t}^i = \sum_{n=k-s+1}^k (x_t^i - y_n^i)^2,
\ee
and $x_t^i$ and $y_n^i$ are the $i$th dimensions of the observation $\vx_t$ and its $n$th nearest neighbor $\vy_n$. $\delta_t^i$ is the contribution of $i$th dimension of the observation $\vx_t$ at time $t$ to the detection statistic. Therefore, by analyzing $\delta_t^i$ for each dimension $i$ during the final increase period of $\Delta_t$, which causes the anomaly alarm, we can identify the dimensions in which anomaly has been observed. To this end, we propose to use a recent history of $\cQ_i = \{\delta_q^i: ~q=\hat{\tau}+1,\ldots,\hat{\tau}+S, ~\forall i\}$ since the last time $\Delta_q = 0$. This time $\hat{\tau}$, the most recent time instance when the detection statistic was zero, can be seen as an estimate of the anomaly onset time. Finally, we apply a $t$-test on the $S$ samples in $\cQ$ to decide whether each dimension $i$ is anomalous.

In particular, we propose the following anomaly localization procedure after the alarm is raised at time $T$:
\begin{enumerate}
	\item Find $\hat{\tau} = \max\{t < T: ~ \Delta_t = 0 \}$
	\item Compute the sample mean and sample standard deviation of $\cQ_i$ for each dimension $i$:
	\begin{align} \label{eq:delta_i}
	& \overline{\delta}_i = \frac{1}{S} \sum_{t = \hat{\tau}+1}^{\hat{\tau}+S} \delta_t^i ~~\text{and}~~ \eta_i = \sqrt{\frac{1}{S-1} \sum_{t = \hat{\tau}+1}^{\hat{\tau}+S} (\delta_t^i-\overline{\delta}_i)^2}
	\end{align}
	\item Identify the anomalous dimensions by applying a $t$-test: 
	\be
	\text{if}~~ \frac{\overline{\delta}_i-\mu_i}{\eta_i/\sqrt{S}} \ge \theta, ~~\text{then dimension $i$ is anomalous},
	\ee 
	where $\mu_i$ is the sample mean of nominal training $\{\delta_1^i,\ldots,\delta_{N_1}^i\}$ values, and $\theta$ is the $(1-\beta)$th percentile, for significance level $\beta$, of Student's $t$-distribution with $S-1$ degrees of freedom. 
\end{enumerate}
The significance level $\beta$, for which a typical value is $0.05$, controls a balance between sensitivity to anomalies and robustness to nominal outliers. For given $\beta$ and $S$ values, the threshold $\theta$ can be easily found from a lookup table for Student's $t$-distribution (e.g., $\theta=6.314$ for $\beta=0.05$ and $S=2$). The number of samples $S$ needs to be at least $2$ to have a degree of freedom at least $1$. In practice, $t$-test is commonly used for small sample sizes, therefore $S$ does not need to be large. Indeed, larger $S$ would cause longer reaction time since the localization analysis would be performed at time $\hat{\tau}+S$, which could be greater than the detection time $T$, incurring extra delay for localization and reaction after detection. 

Localization by ODIT-2 is slightly different. Since $\log L_{(K_N)}$ and $\log L'_{(K_M)}$ are constant, the increase that causes the alarm takes place in $\log L_t - \log L'_t$. Writing $L_t$ and $L'_t$ in terms of the contributions from the $d$ dimensions, $\delta_t^{i}$ and $\delta_t^{'i}$, respectively, as in \eqref{eq:L_contr}, the increase in the difference $(\delta_t^{i} - \delta_t^{'i})$ for some $i$ leads to the increase in the decision statistics. 
Similar to ODIT, firstly $\hat{\tau}$ is found after a detection. Then, in the second step, the $\overline{\delta}^i$ and $\eta_i$ are computed by replacing $\delta_t^i$ with $(\delta_t^{i} - \delta_t^{'i})$ in \eqref{eq:delta_i}. Finally, in the third step, $\mu_i$ corresponds to the sample mean of nominal training $\{(\delta_1^{i} - \delta_1^{'i}),\ldots,(\delta_{N_1}^{i} - \delta_{N_1}^{'i})\}$ values.

\section{DDoS Attack Mitigation} 
\label{sec:ddos}

Distributed Denial-of-Service (DDoS) attack is a major security problem in today's widely-networked systems and requires effective solution approaches \cite{DDoSDouligeris,kolias2017ddos}. DDoS attack is traditionally known as a type of cyber-attack targeting an Internet service, with the intention of making it unavailable for the legitimate users. Nevertheless, it has also been recently investigated in the cyber-physical systems domain, such as the smart grid \cite{soltan2018blackiot}. DDoS attack is typically performed by overwhelming the target with malicious requests from multiple geographically distributed sources. The attacker first builds a network of malicious devices known as ``botnet" by infecting them with malware, and then remotely controls these devices to synchronously send some form of service requests to the target, which initiates a DDoS attack. The size of botnet both in the number of compromised devices and geographical distribution determines the threat level of a DDoS attack. It is extremely difficult to successfully mitigate a large-scale DDoS attack centrally at the attacked site without disrupting the regular service to legitimate users, as recently demonstrated by the massive DDoS attacks empowered by Internet-of-things (IoT) devices \cite{kolias2017ddos}.

\emph{Low-Rate DDoS:} The proliferation of IoT devices exacerbates the DDoS attack problem as many IoT devices, such as Internet-connected sensors, have low security measures, making them vulnerable to malware infections \cite{IoTDataset}. Abundance of low-security IoT devices worldwide enables an even more challenging new type of DDoS attack, called low-rate DDoS \cite{newinfobased}, which is considered a stealth attack since the amount of anomalous service requests from each compromised device can be quite low. Such low-rate change in the device behavior can easily bypass local intrusion detection systems (IDSs) that rely on observing raw data, such as data filters and firewalls. Yet, a synchronous low-rate DDoS attack from huge number of compromised devices, e.g., millions of IoT devices, can easily cause an overwhelming aggregated service request, and thus the failure of target. 
Successful DDoS attack mitigation requires quick detection of attack, and accurate identification of sources of malicious requests so that appropriate countermeasures can be taken against the attack. The timely detection of low-rate DDoS attacks is quite challenging at the local level, e.g., at the routers close to IoT devices. Although detection is trivial at the target due to the overwhelming aggregated service requests, accurate identification of attacking nodes and as a result mitigation of the DDoS attack in a centralized fashion is not tractable. 

\emph{Challenges:} There are several challenges for mitigating low-rate DDoS attacks. 
\emph{(i) High-dimensionality:} 
DDoS attacks inherently relate to large-scale systems. 
Therefore, the proposed methods need to scale well to large systems. Particularly, for low-rate DDoS attacks, timely and accurate detection at a local level is challenging due to the similarity of attack behavior to the nominal behavior. Multivariate anomaly detection techniques can greatly facilitate timely and accurate detection, however even in a local IoT network, dimensionality, i.e., number of devices, makes joint probability density estimation intractable for parametric methods. 
\emph{(ii) Heterogeneity:} 
The heterogeneous nature of IoT results in complex probability distributions even under nominal settings. Each device type in the network has different usage characteristics. For instance, it is expected for computer, phone, smart watch and temperature sensor in a network to have different operational baselines. Furthermore, even the nominal probability distribution of a single device is usually complicated due to its different operation modes, such as active use, passive use at the background, and hibernation. 
\emph{(iii) Unknown attack types:} 
Due to the myriad of vulnerabilities in a network of low-security IoT devices, it is not possible to know future attack patterns. The conventional signature-based IDSs are not effective since they can only detect a predefined set of attack patterns. For the same reason, parametric detection techniques which assume probabilistic models for anomalies are not feasible as well. To be able to detect unknown anomaly types, a nonparametric detection method is needed.

\emph{Application of ODIT:} Considering the challenges mentioned above ODIT provides an effective local DDoS attack mitigation approach that can handle high-dimensionality, heterogeneity, and unknown attack types for quick and accurate detection. Utilizing the hierarchical structure of large-scale systems, such as the Internet and the power distribution network, multiple ODITs running at local level, such as routers and data aggregators, can provide a complete IDS for DDoS attack mitigation.
Since ODIT is a generic anomaly detection method, we do not specify the observed data type, i.e., service request, in the following simulations for DDoS mitigation. For instance, following the commonly used DDoS concept in computer networks (e.g., flooding-based DDoS \cite{DDoSDouligeris}) the observed data vector could be the number of packets in unit time, such as packets per second, from a number of devices in the network \footnote{The real-time demonstration of ODIT using an IoT testbed can be seen at \url{https://youtu.be/zQexZgB5AMs}.}; or considering a power delivery network like in \cite{soltan2018blackiot} the observed data dimensions could be the power demand from houses.

\subsection{Compared Methods}

We compare the performance of the proposed methods with two state-of-the-art detection methods for DDoS attacks. The information metric-based method \cite{newinfobased}, and the deep autoencoder method \cite{IoTDataset} are used for comparison in the simulation (Section \ref{sec:simulated_exp}) and real dataset (Section \ref{sec:NBaIoT}) experiments, respectively. The latter was proposed in the paper that presented the N-BaIoT dataset \cite{IoTDataset}, thus we use it to evaluate the performance of the proposed ODIT detectors on this dataset in Section \ref{sec:NBaIoT}. The former is a window-based method that assumes Gaussian distribution for the nominal data, and Poisson distribution for the attack data. Specifically, in the training phase, it fits a Gaussian distribution to a nominal dataset, and then in the test phase it fits a Poisson distribution to a window of samples. By sliding the window and updating the Poisson distribution at each time it computes its detection statistic as 
\begin{align}
& D_{\alpha}(P,Q) = D_{\alpha}(P||Q) + D_{\alpha}(Q||P) 
\end{align}
where $P$ and $Q$ are the estimated Gaussian and Poisson distributions, and $D_{\alpha}(P||Q)$ is the R\'{e}nyi divergence between $P$ and $Q$ with parameter $\alpha \in (0,\infty)$. Since this method is a window-based, its performance is highly dependent on the choice of the window size. For small window size, the accuracy of the probability distributions would not be good, resulting in poor performance, while large window size would increase the detection delay, as the attacks can only be detected at the end of the initial window. In the worst scenario assuming that the window size is $W$ and anomaly starts at the beginning of the window, the detection delay would be at least $W$. Moreover, for large window size, it would take more time to see the effect of attack in the estimated Poisson distribution, and thus longer detection delays. This method is designed to capture the increase in the average data rate with respect to the average in the training dataset.
We also compare the proposed methods with the conventional data filtering method that filters out the service requests, in particular data packets, from nodes whose number in a certain period (e.g., packet rate) exceeds a predefined threshold.

\subsection{Experiment on Simulated Data}
\label{sec:simulated_exp}

In the first experiment, as part of a low-rate DDoS attack scenario, we simulated an IoT network with $d = 50$ devices of different types, each having different nominal data transmission rates. Although the N-BaIoT dataset used in Section \ref{sec:NBaIoT} is also collected from a similar IoT network, the attack magnitudes (i.e., increase in the data rates) are significantly higher than what we consider as low-rate DDoS here. We perform this simulation study to investigate a low-rate DDoS attack scenario in larger IoT networks.
For example, the nominal data rate of a temperature sensor is considerably lower than that of a surveillance camera or a computer. In this simulation setup, $30\%$ of the devices have two modes of operation, active and inactive states, with higher data rates in the former. The rest of the devices have a single baseline representing the background traffic in practical networks. The data rates of each device are generated independently from each other from a Gaussian distribution. For a device, data rates over time are independent and identically distributed. The mean data rates are chosen randomly in $[10,50]$ for inactive states,  in $[50,90]$ for active states, and in $[10,100]$ for the devices with single states. The same variance $\sigma^2 = 5$ is used for all devices. Note that data rates of the bimodal devices with active and inactive states follow a mixture of two Gaussian distributions. The frequency of active and inactive states are set to be equal.
Assume that an attacker initiates a DDoS attack at time $\tau = 101$ through several compromised devices present in the network. When an attack starts, the compromised devices start sending data at a higher rate with a $5$ standard deviation increase.

\begin{figure}[t]
\begin{center}
	
	\includegraphics[width=\linewidth]{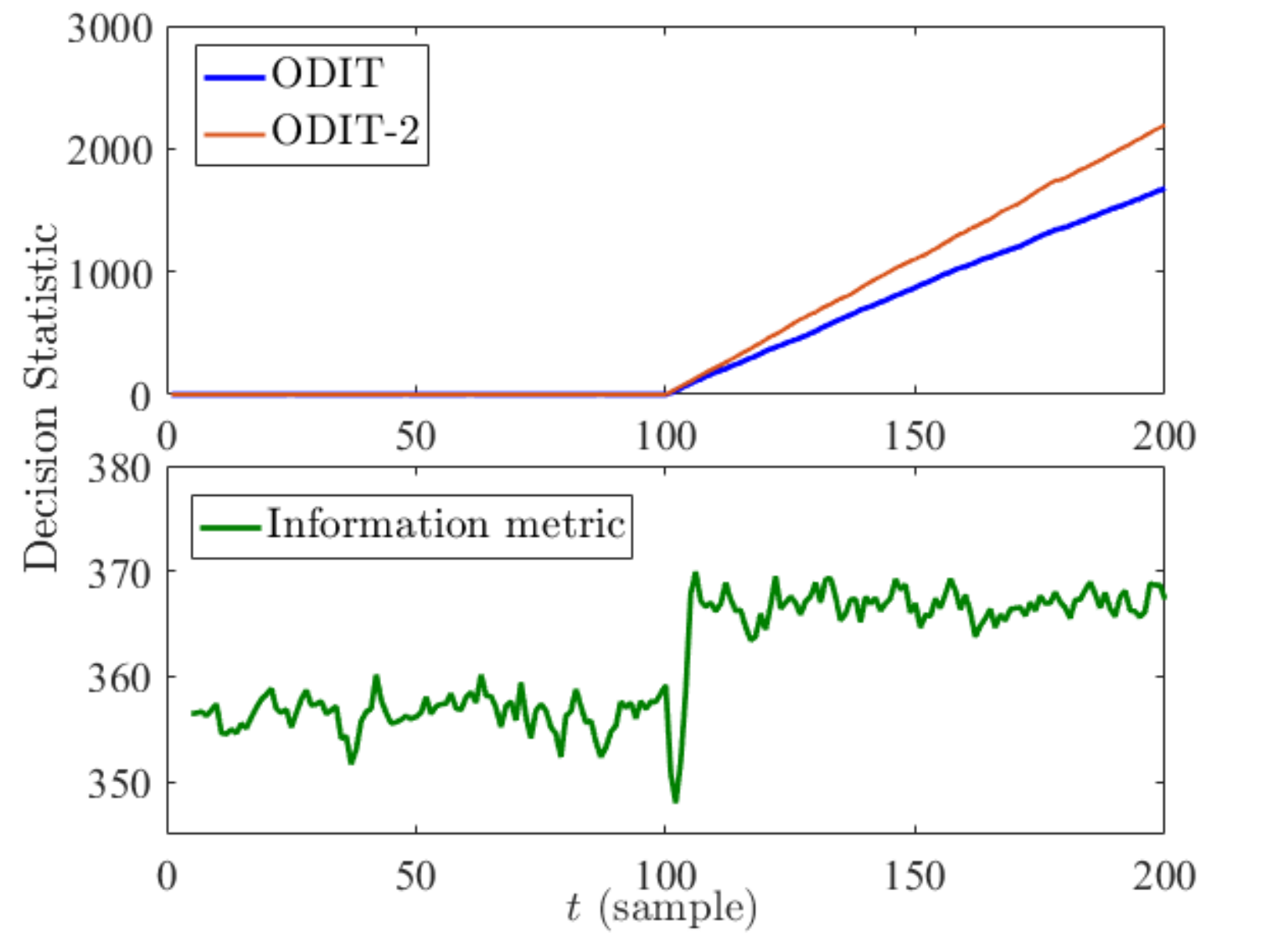}

	\caption{\label{fig:stats2} Decision statistics of ODIT, ODIT-2 and the information metric-based method in the DDoS simulation study.}
	
\end{center}

\vspace{-0.2in}
\end{figure}

\begin{figure}[t]
\begin{center}
	
	\includegraphics[width=\linewidth]{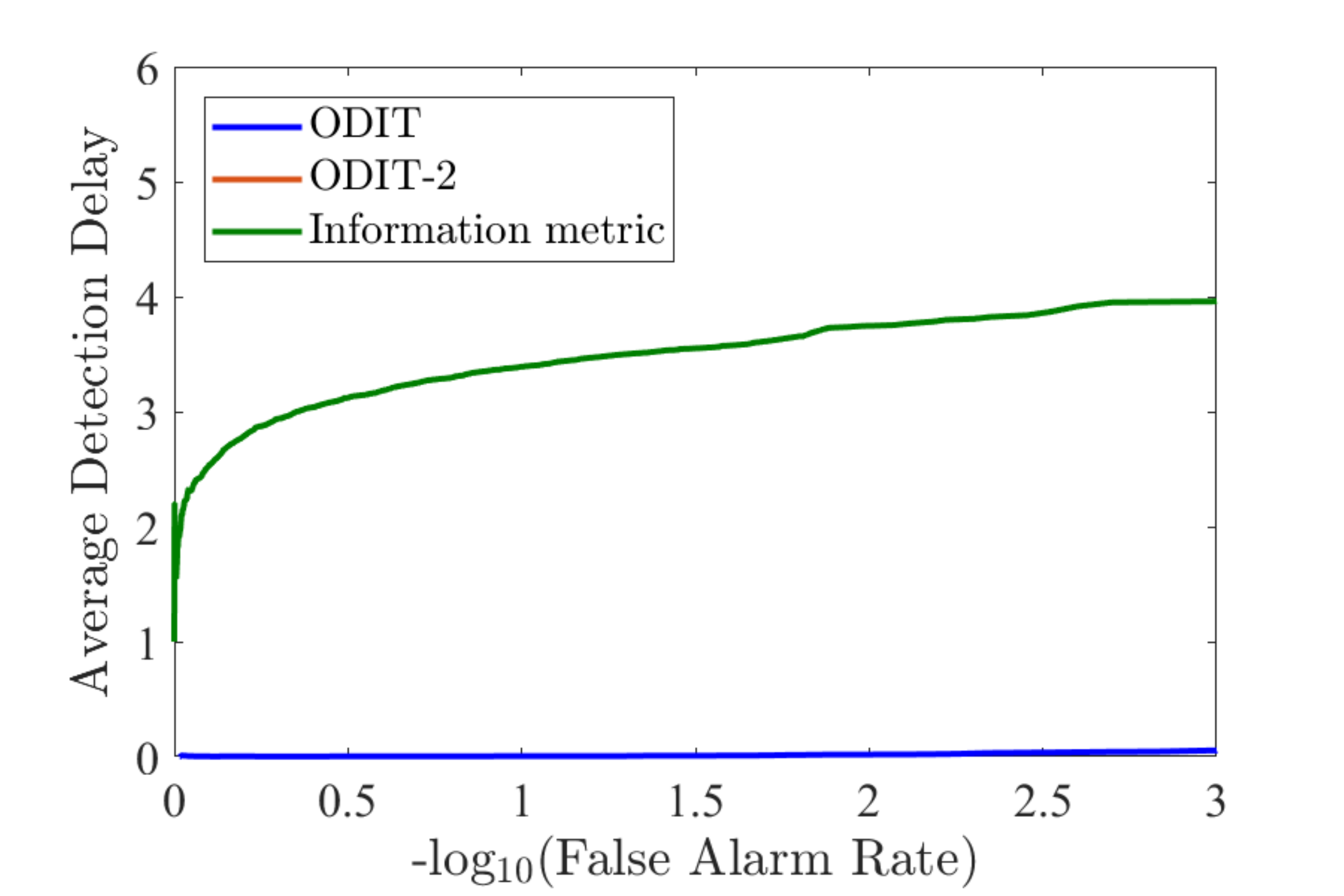}

	\caption{\label{fig:ds2} Average detection performance of the proposed ODIT, ODIT-2 and information metric-based detectors in the DDoS simulation study.}
	
\end{center}

\vspace{-0.2in}
\end{figure}

In the ODIT algorithms, we set the parameters as $k=1$, $s=1$, $\alpha_1 = 0.05$, $\alpha_2 = 0.05$, $\gamma=1$, $S=2$. The results are obtained using $N=2\times 10^5$ nominal instances and $M=10^5$ anomalous instances.
Fig. \ref{fig:stats2} shows the decision statistics of ODIT, ODIT-2 and the information metric-based algorithm proposed in \cite{newinfobased}. As depicted in the figure, ODIT statistics exhibit an abrupt increase. The best window size for the information metric-based method is found to be $W = 5$. The information distance starts increasing only when the window contains enough number of anomalous data instances. This result is consistent with the average performance results (average detection delay vs. false alarm rate) shown in Fig. \ref{fig:ds2}. The ODIT-2 detector achieves zero detection delay with no false alarm. Similarly, ODIT achieves very small average detection delay while satisfying very low false alarm rate at $10^{-3}$.
Although the information metric-based method also achieves reasonable detection delays, compared to ODITs it suffers from its window-based nature. The smaller or larger window sizes do not give better results due to insufficient accuracy in probability distribution estimations and less sensitivity to anomalies, respectively.


\begin{figure}[t]
\begin{center}
	
	\includegraphics[width=\linewidth]{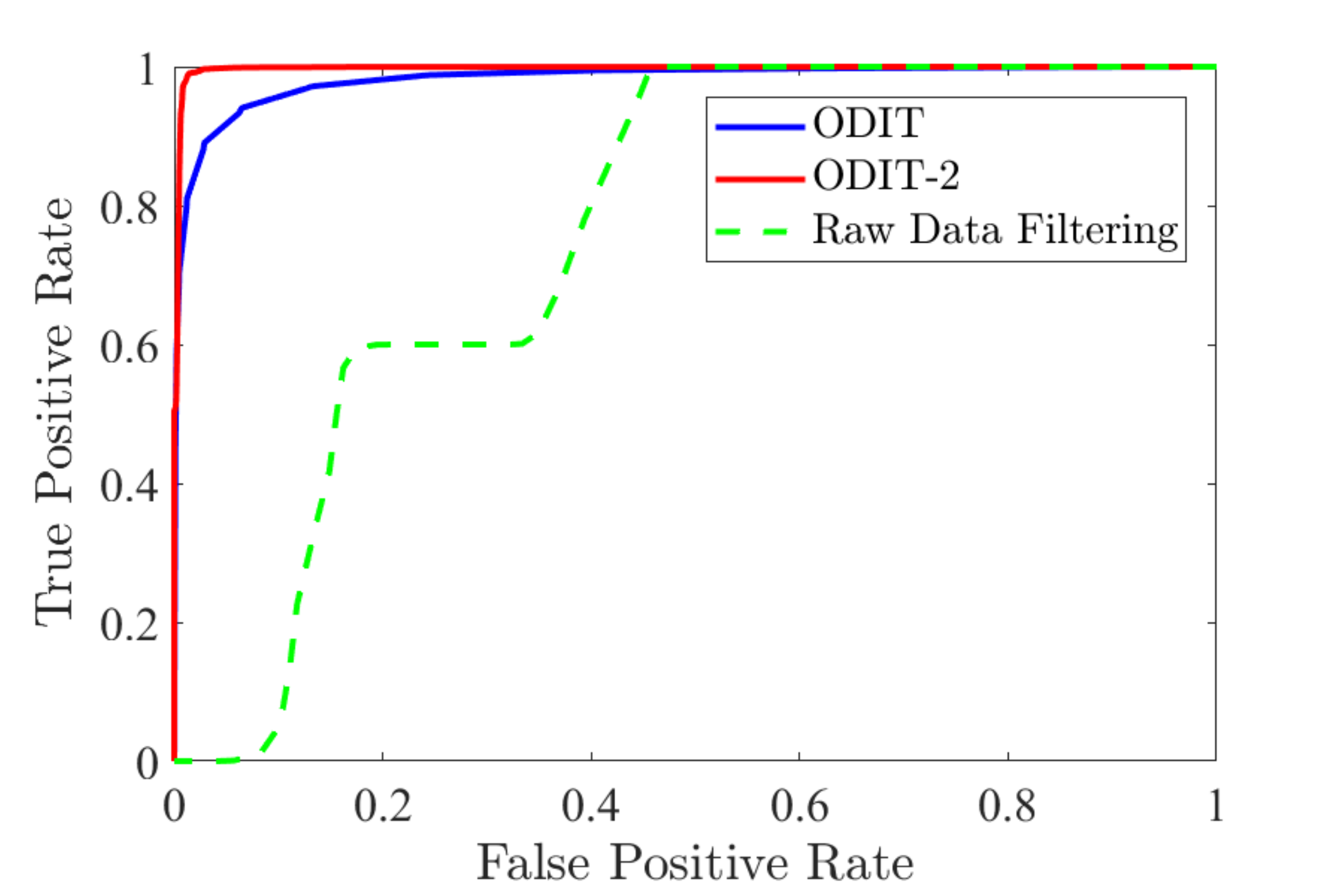}

	\caption{\label{fig:roc} ROC curve for identifying malicious devices in DDoS simulations.}
	
\end{center}

\vspace{-0.2in}
\end{figure}

The receiver operator characteristic (ROC) curves for localization of the malicious devices are shown in Fig. \ref{fig:roc}. In comparison with the data filtering approach, ODIT and ODIT-2 successfully identify the malicious devices with probability $0.95$ and $1$, respectively, while satisfying the false positive rate of $0.05$. The conventional data filtering approach identifies a device as anomalous if its data rate exceeds a predefined threshold. Due to the small attack magnitudes in the simulated low-rate DDoS attack the data filtering approach fails to achieve high identification probability while satisfying small false positive rates. 

\subsection{Experiment on a Real Dataset: N-BaIoT}
\label{sec:NBaIoT}

In the second experiment, we evaluated the proposed ODIT algorithms using the N-BaIoT dataset, which consists of real IoT data traffic observations including botnet attacks. This data is collected from $9$ IoT devices including doorbell, thermostat, baby monitor, etc. infected by the Mirai and BASHLITE malware \cite{IoTDataset},\cite{IoTDataset2}. Here we only consider the Mirai attack dataset. 
The benign and attack datasets for each device is composed of $115$ features summarizing traffic statistics over different temporal windows. The dataset is collected for each device separately, and lacks timestamp. The number of instances is varied for each device and attack type. Therefore, we formed the training and test sets by randomly choosing data instances from each device. To form a network-wide instance for multivariate detection we stack the chosen instances from $9$ devices into a single vector of $1035$ dimensions. This way, we obtain a nominal training set with $N=10,000$ instances. We also build an anomalous training set with $M=5,000$ instances for the Ecobee thermostat device (device 3). To test for both known and unknown attack types we let ODIT-2 train only on attack data from device 3, and test under two scenarios: 
\begin {enumerate*} [(i) ]%
\item device 3 (Ecobee Thermostat) is compromised (known anomaly type) \item device 6 (Provision PT-838 security camera) is compromised (unknown anomaly type).
\end {enumerate*}
We form the test data similarly to the training data, assuming that the respective device gets compromised and starts sending malicious traffic at $t = 101$. 
In the ODIT algorithms we set parameters as $k=s=\gamma=1$, $\alpha_1 = 0.05$, $\alpha_2 = 0.1$, $S=2$.

\begin{figure}[t]
\begin{center}
	
	\includegraphics[width=\linewidth]{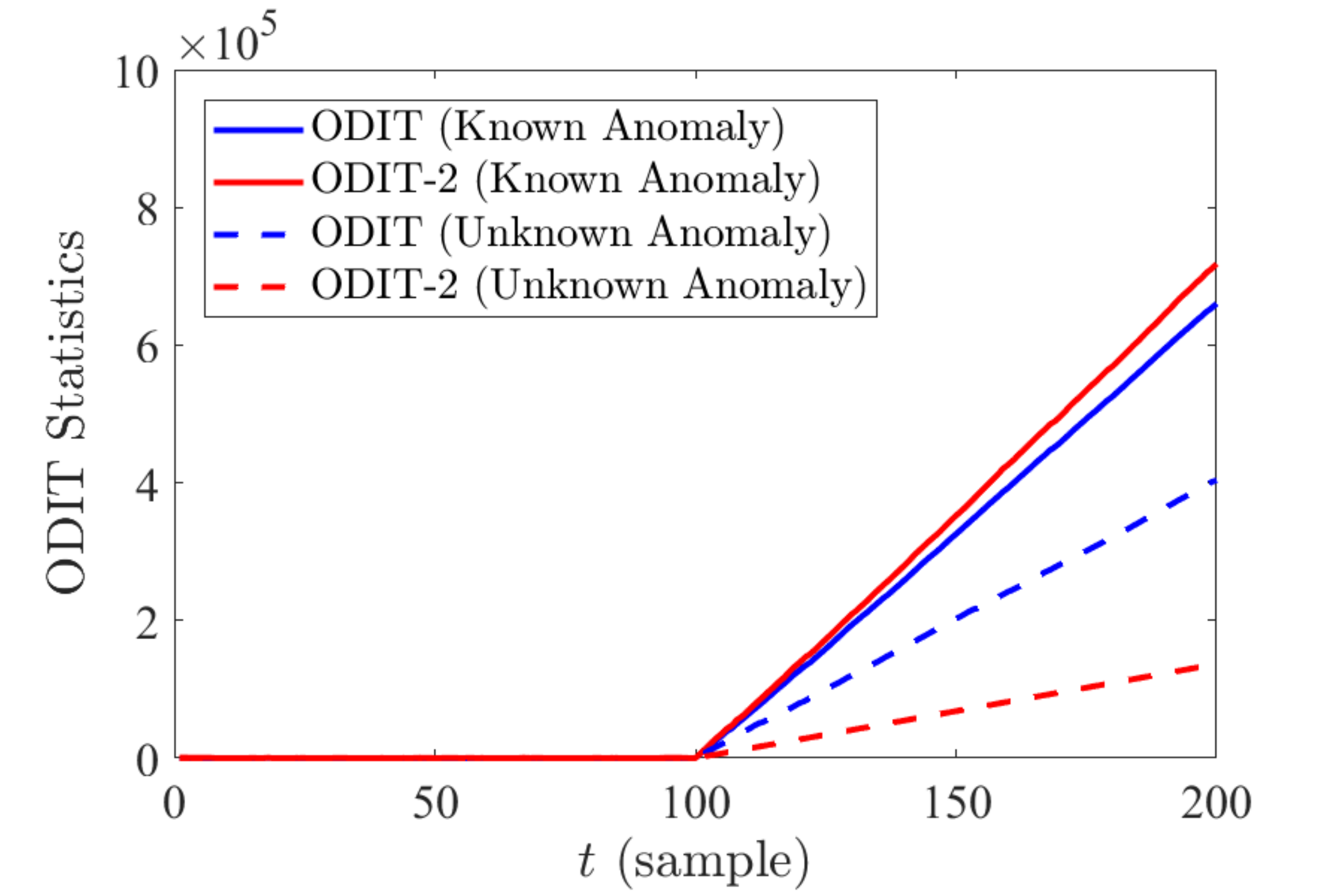}

	\caption{\label{fig:iot_ds} Desision statistics of ODIT and ODIT-2 in both known and unknown attack scenarios for the N-BaIoT dataset.}
	
\end{center}

\vspace{-0.2in}
\end{figure}

\begin{figure}[t]
\begin{center}
	
	\includegraphics[width=\linewidth]{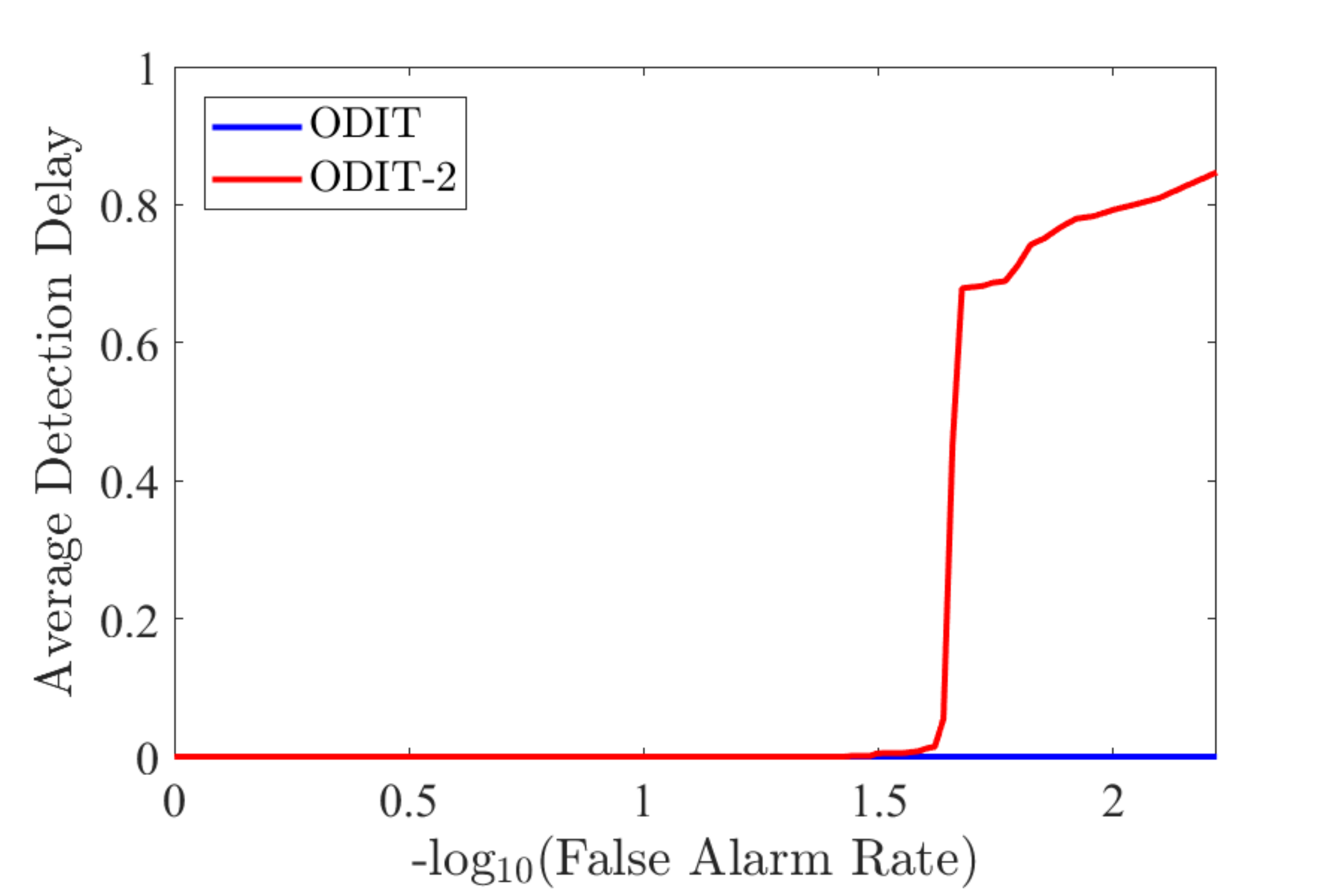}

	\caption{\label{fig:iot_performance} Performance comparison for ODIT and ODIT-2 in the unknown attack scenario for the N-BaIoT dataset.}
	
\end{center}

\vspace{-0.2in}
\end{figure}

An example of the decision statistics for ODIT and ODIT-2 under the two scenarios are shown in Fig. \ref{fig:iot_ds}. ODIT is able to detect the attack with zero detection delay and zero false alarm in all trials in both known and unknown attack scenarios (Fig. \ref{fig:iot_performance}). 
As for ODIT-2, which trains also on attack data from device 3, in the known attack scenario, zero detection delay with zero false alarm in all trials is achieved, similar to ODIT. 
Fig. \ref{fig:iot_ds} shows that the ODIT-2 decision statistic steadily rise even for the unknown attack when device 6 is attacking, yet with a smaller slope than that of ODIT, as expeected. However, such a rise is not guaranteed to happen in general for unknown anomaly types. When an unknown anomaly occurs in the test observations, depending on whether the anomalous observations are relatively similar to the nominal dataset or to the anomalous dataset, ODIT-2 may or may not detect the anomaly. In the case where the anomalous data instances are relatively more similar to the nominal set than to the anomaly set, ODIT-2 statistics will remain zero and it will fail in detecting the anomaly. In the experiment, however, the unknown anomaly type, the attack data from device 6, is relatively more similar to the anomaly training set, the attack data from device 3. Therefore, ODIT-2 is able to detect it, as shown in Fig. \ref{fig:iot_performance}, where the average detection performances of ODIT and ODIT-2 are given for scenario 2 (unknown anomaly).

\begin{figure}[t]
\begin{center}
	
	\includegraphics[width=\linewidth]{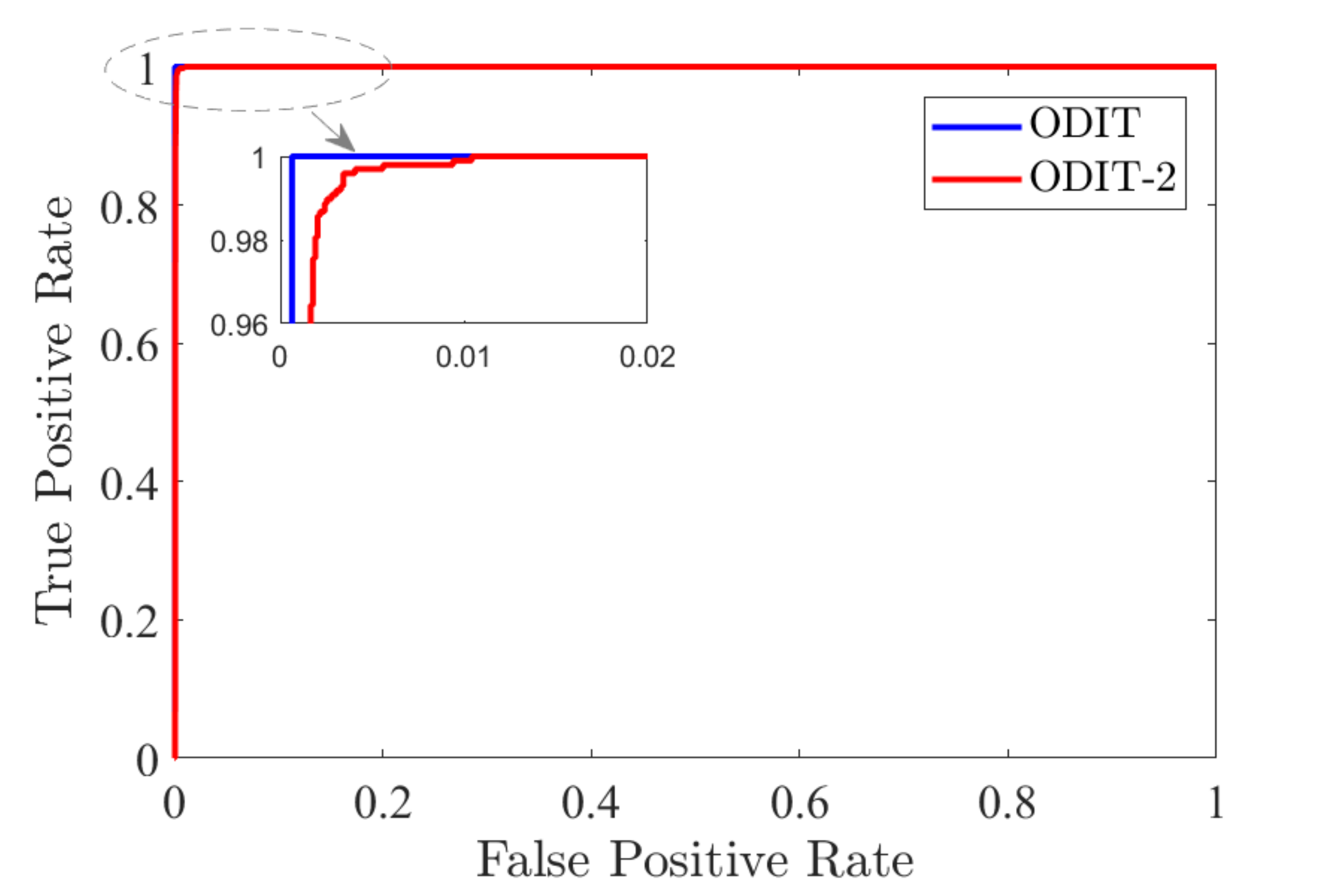}

	\caption{\label{fig:iot_Roc} ROC curve for anomaly localization using ODIT and ODIT-2 in the known attack scenario for the N-BaIoT dataset.}
	
\end{center}

\vspace{-0.2in}
\end{figure}

Next, the identification of malicious device is investigated in Fig. \ref{fig:iot_Roc} in terms of the ROC curve (true positive rate vs. false positive rate) under the known anomaly scenario. Both variations of our proposed method identify the malicious device with very high probability while achieving small false alarm rates such as $0.01$. We calculate the contribution of each device to the decision statistic in \eqref{eq:delta_i} as the sum of the contributions of all 115 dimensions corresponding to the device.

\begin{figure}[t]
\begin{center}
	
	\includegraphics[width=\linewidth]{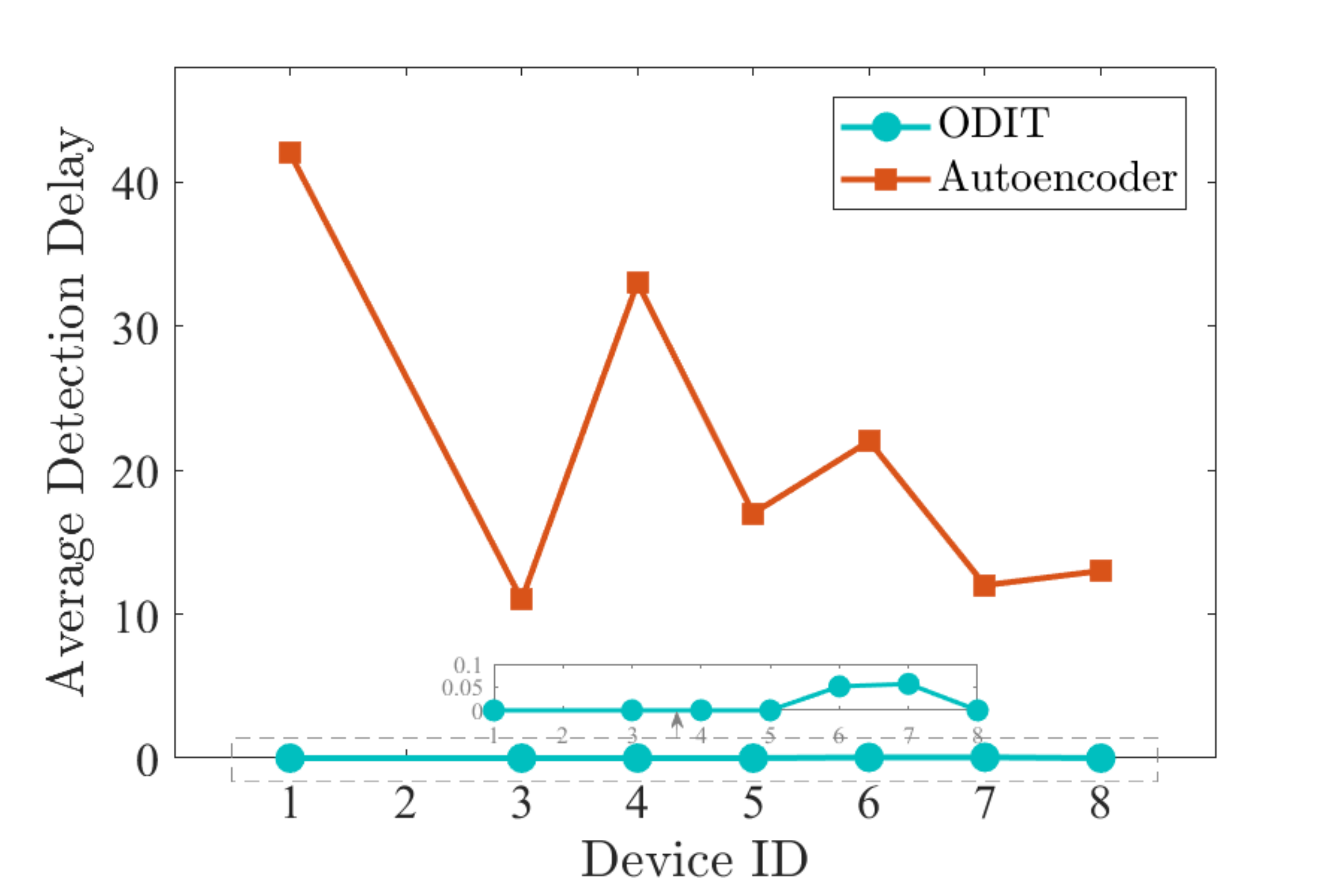}

	\caption{\label{fig:autoencoder_comp_delay} Average detection delay of the autoencoder method \cite{IoTDataset} and ODIT in terms of number of samples for each device attack scenario.}
	
\end{center}

\vspace{-0.2in}
\end{figure}

\begin{figure}[t]
\begin{center}
	
	\includegraphics[width=\linewidth]{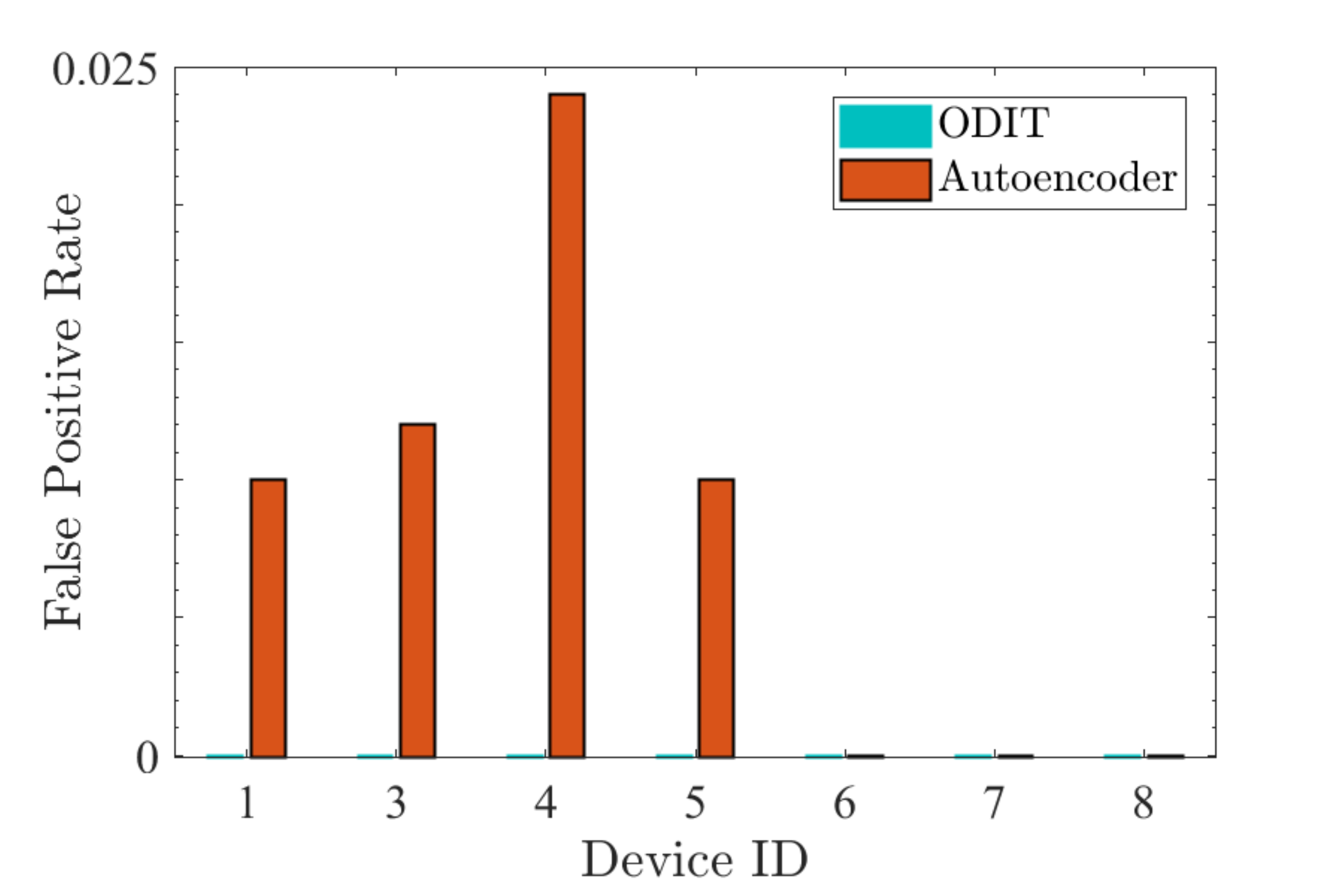}

	\caption{\label{fig:autoencoder_comp_FPR} False postitive rate of the autoencoder method \cite{IoTDataset} and ODIT for each device attack scenario.}
	
\end{center}

\vspace{-0.2in}
\end{figure}

We also compare the performance of ODIT to the deep autoencoder-based detection method \cite{IoTDataset}, as they both train only on the nominal data. The autoencoder method marks each observation instance as nominal or anomalous, and employs majority voting on a moving window of size $ws^*$ (to control the false positive rate), raising alarm only if the majority of the instances within the window are marked as anomalous. Due to its window-based majority rule, the sample detection delay (i..e., the number of anomalous instances observed before the detection) is at least $\lfloor \frac{ws^*}{2} \rfloor + 1$. Whereas, the sequential nature of ODIT enables immediate detection together with zero false alarm, as demonstrated in Fig. \ref{fig:autoencoder_comp_delay} and Fig. \ref{fig:autoencoder_comp_FPR}. Following the analysis in \cite{IoTDataset} for each device, the sample detection delay and the false positive rate of both methods are compared in Fig. \ref{fig:autoencoder_comp_delay} and Fig. \ref{fig:autoencoder_comp_FPR}, respectively. 
The optimum window sizes reported in \cite{IoTDataset} for each device are used for the autoencoder method.

\subsection{Online Learning Scheme: ODIT-uni}

In this section, we present experiment results to demonstrate the practical advantage of the unified framework ODIT-uni, proposed in Section \ref{sec:ODIT-uni}. Following the simulated and real-data experiments of Sections \ref{sec:simulated_exp} and \ref{sec:NBaIoT} we train the algorithms on the nominal data and anomaly data for a specific attack type. For the N-BaIoT dataset, we repeat the scenario 2 test, in which device 6 (Provision PT-838 security camera) starts sending malicious traffic while only attack data from device 3 is used to train ODIT-2. 
We extend the simulation experiment of Section \ref{sec:simulated_exp} by testing the trained algorithms on a new anomaly type. 
Specifically, at time $t=101$ a different set of devices start acting maliciously. 

\begin{figure}[t]
\begin{center}
	
	\includegraphics[width=\linewidth]{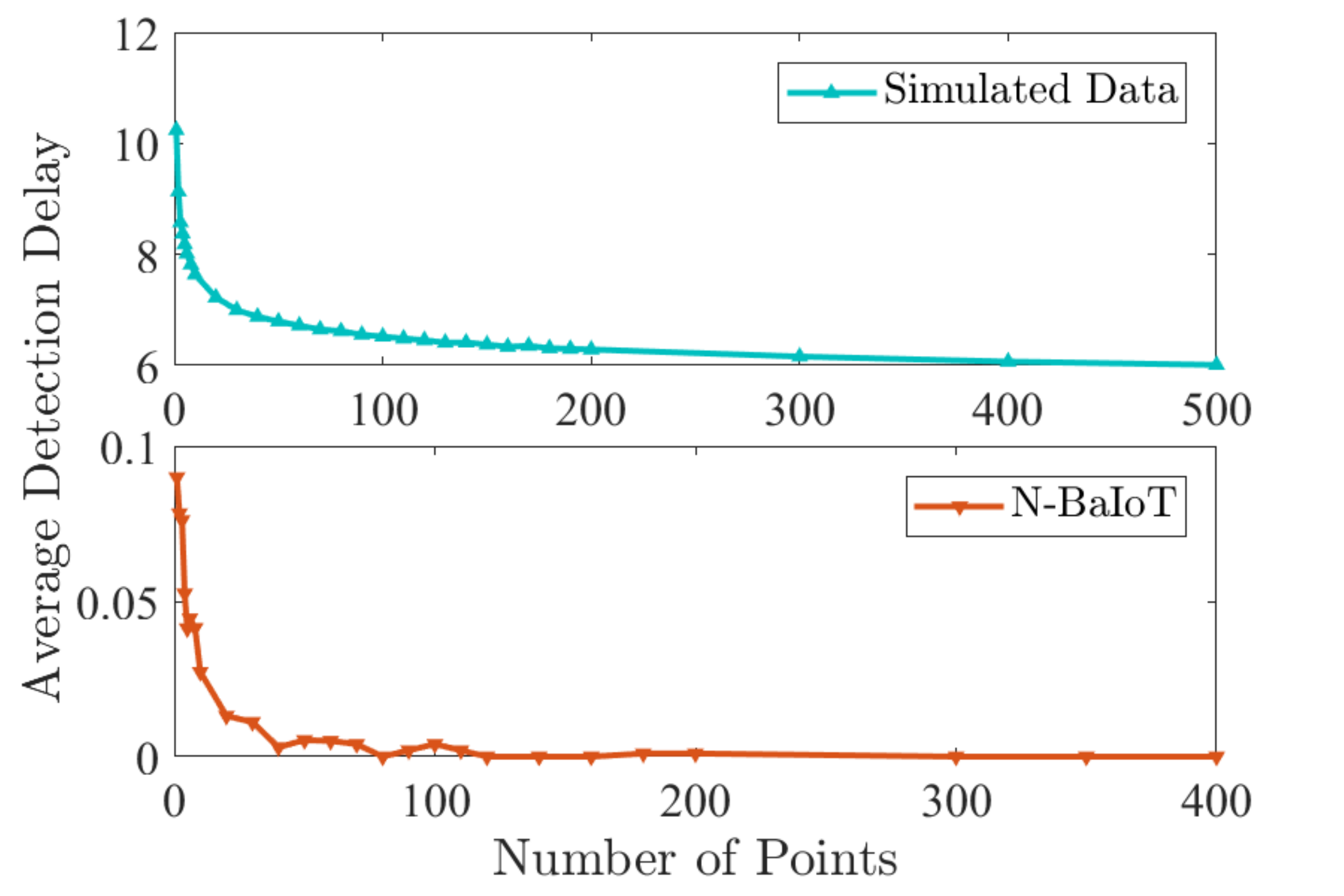}

	\caption{\label{fig:unified_framework} Average detection delay of ODIT-2 vs. number of the new-anomaly instances added to the training set for simulation and real-data experiments. }
	
\end{center}

\vspace{-0.2in}
\end{figure}

Fig. \ref{fig:unified_framework} shows, for both the simulated and N-BaIoT datasets, the average detection delay by ODIT-2 for a constant false alarm rate of $0.01$, versus the number of the data points from the new anomaly type added to the anomaly training set. In both cases, as the number of the confirmed instances added to the anomaly training set grows, ODIT-2 detection delay decreases. The confirmation can be through either a human expert or a sufficiently high decision threshold for ODIT which avoids false alarms, as explained in Section \ref{sec:ODIT-uni}. In the simulated data, ODIT-2 is not able to detect the new anomaly type at the beginning without seeing any representative instance. However, even after seeing only a single instance of the new anomaly type, it is able to detect it with a reasonable delay around $10$. Whereas, in the N-BaIoT dataset, ODIT-2 is able to detect the unknown anomaly at the first encounter with an average delay of $0.79$, and the average delay converges to zero as the training set is enhanced with instances from the new anomaly type. 
In this way, ODIT-uni detects the unknown anomaly types through ODIT, and over time learns the geometry of new anomalies and improves its detection performance through ODIT-2.

\ignore{
Next, we present an experiment to evaluate the average performance of the unified framework for the unknown anomalies in comparison to where the original ODIT algorithms running individually without learning scheme. We Follow the trained models mentioned above. In this experiment we consider the scenario where only the novel anomaly types occur in the observations (the anomaly types not used for the initial training of ODIT-2). The threshold of ODIT is set sufficiently high to avoid false alarms by ODIT, and consequently false inclusions of nominal data instances to the anomaly train set. We test the unified framework with different thresholds to evaluate its average performance in terms of the false alarm rate and average detection delay of the unified framework (number of the samples observed before either ODIT or ODIT-2 in unified framework detects an anomaly first). For each pair of predefined thresholds $h_1$ and $h_2$, we performed test on set of novel anomaly types for $500$ trials. It is expected that in the first trials the $\Delta_t^{(1)}$ will detect the anomaly, resulting in the anomaly instances being included in the anomaly train set, whereas in the later trials, the ODIT-2 part of the unified framework is expected to detect anomaly with lower detection delay. As the anomaly train set grows to contain more and more new anomaly instances, the detection by ODIT-2 part of the framework improves resulting in the improvement in the average performance of the unified framework. Fig. \ref{fig:unif_average} corroborates the improvement on the average performance of unified framework as compared to the individual ODIT algorithms in non-learning manner.

\begin{figure}[t]
\vspace {-0.1in}
\begin{center}	
	\includegraphics[width=\linewidth]{Unified/performance_compare_to_unified.png}
	\caption{\label{fig:unif_average} The comparison of average performance of Unified Framework vs. ODIT and ODIT-2 running in individual for the unknown anomaly type. }	
\end{center}
\vspace{-0.2in}
\end{figure}

}

\section{Conclusion} \label{sec:conc}

In this paper, we proposed an algorithm, called ODIT, that is suitable for quick and accurate anomaly detection and localization in high dimensional systems which require multivariate (i.e., joint) monitoring of system components. 
Our proposed anomaly detection method is generic and applicable to various contexts as it does not assume specific data types, probability distributions, and anomaly types. It only requires a nominal training set, and achieves asymptotic optimality in terms of minimizing average detection delay for a given false alarm constraint. We also showed how to benefit from available anomalous data (ODIT-2), and presented an online learning scheme (ODIT-uni) that detects unknown anomaly types and over time improves its performance by learning from detected anomalies. 
We evaluated the performance of our method in the context of DDoS attack detection and botnet detection using a simulated dataset and a real dataset. The experiments verified the advantage of proposed online learning method, and also showed that the proposed ODIT methods significantly outperform the state-of-the-art anomaly/change detection methods in terms of average detection delay and false alarm rate. 

The proposed algorithms assume static nominal behavior and a static set of data dimensions. For instance, the proposed online learning scheme updates its anomaly knowledge in real-time, but it does not update its nominal data repository. 
Extending it to dynamic settings, such as an IoT network with dynamic topology and changing nominal behavior, remains to be an important future research direction.

\bibliographystyle{IEEEtran}
\bibliography{ref}

\begin{thebibliography}{10}
\providecommand{\url}[1]{#1}
\csname url@samestyle\endcsname
\providecommand{\newblock}{\relax}
\providecommand{\bibinfo}[2]{#2}
\providecommand{\BIBentrySTDinterwordspacing}{\spaceskip=0pt\relax}
\providecommand{\BIBentryALTinterwordstretchfactor}{4}
\providecommand{\BIBentryALTinterwordspacing}{\spaceskip=\fontdimen2\font plus
\BIBentryALTinterwordstretchfactor\fontdimen3\font minus
  \fontdimen4\font\relax}
\providecommand{\BIBforeignlanguage}[2]{{%
\expandafter\ifx\csname l@#1\endcsname\relax
\typeout{** WARNING: IEEEtran.bst: No hyphenation pattern has been}%
\typeout{** loaded for the language `#1'. Using the pattern for}%
\typeout{** the default language instead.}%
\else
\language=\csname l@#1\endcsname
\fi
#2}}
\providecommand{\BIBdecl}{\relax}
\BIBdecl

\bibitem{SURVEY}
V.~Chandola, A.~Banerjee, and V.~Kumar, ``Anomaly detection: A survey,''
  \emph{ACM computing surveys (CSUR)}, vol.~41, no.~3, p.~15, 2009.

\bibitem{newinfobased}
Y.~Xiang, K.~Li, and W.~Zhou, ``Low-rate ddos attacks detection and traceback
  by using new information metrics,'' \emph{IEEE transactions on information
  forensics and security}, vol.~6, no.~2, pp. 426--437, 2011.

\bibitem{medicAD}
H.~Zhang, J.~Liu, and N.~Kato, ``Threshold tuning-based wearable sensor fault
  detection for reliable medical monitoring using bayesian network model,''
  \emph{IEEE Systems Journal}, vol.~12, no.~2, pp. 1886--1896, 2018.

\bibitem{BIGDATASURVEY}
R.~A.~A. Habeeb, F.~Nasaruddin, A.~Gani, I.~A.~T. Hashem, E.~Ahmed, and
  M.~Imran, ``Real-time big data processing for anomaly detection: A survey,''
  \emph{International Journal of Information Management}, 2018.

\bibitem{CORR}
V.~Avanesov, N.~Buzun \emph{et~al.}, ``Change-point detection in
  high-dimensional covariance structure,'' \emph{Electronic Journal of
  Statistics}, vol.~12, no.~2, pp. 3254--3294, 2018.

\bibitem{taposh1}
T.~Banerjee, H.~Firouzi, and A.~O. Hero~III, ``Quickest detection for changes
  in maximal knn coherence of random matrices,'' \emph{arXiv preprint
  arXiv:1508.04720}, 2015.

\bibitem{soltan2018blackiot}
S.~Soltan, P.~Mittal, and H.~V. Poor, ``Blackiot: Iot botnet of high wattage
  devices can disrupt the power grid,'' in \emph{27th $\{$USENIX$\}$ Security
  Symposium ($\{$USENIX$\}$ Security 18)}, 2018, pp. 15--32.

\bibitem{spacecraft}
K.~Hundman, V.~Constantinou, C.~Laporte, I.~Colwell, and T.~Soderstrom,
  ``Detecting spacecraft anomalies using lstms and nonparametric dynamic
  thresholding,'' \emph{arXiv preprint arXiv:1802.04431}, 2018.

\bibitem{svmbased}
I.~Steinwart, D.~Hush, and C.~Scovel, ``A classification framework for anomaly
  detection,'' \emph{Journal of Machine Learning Research}, vol.~6, no. Feb,
  pp. 211--232, 2005.

\bibitem{c3}
M.~Sedghi, G.~Atia, and M.~Georgiopoulos, ``Kernel coherence pursuit: A
  manifold learning-based outlier detection technique,'' in \emph{2018 52nd
  Asilomar Conference on Signals, Systems, and Computers}.\hskip 1em plus 0.5em
  minus 0.4em\relax IEEE, 2018, pp. 2017--2021.

\bibitem{c4}
------, ``Low-dimensional decomposition of manifolds in presence of outliers,''
  in \emph{2019 IEEE 29th International Workshop on Machine Learning for Signal
  Processing (MLSP)}.\hskip 1em plus 0.5em minus 0.4em\relax IEEE, 2019, pp.
  1--6.

\bibitem{j1}
------, ``Robust manifold learning via conformity pursuit,'' \emph{IEEE Signal
  Processing Letters}, vol.~26, no.~3, pp. 425--429, 2019.

\bibitem{infobased}
W.~Lee and D.~Xiang, ``Information-theoretic measures for anomaly detection,''
  in \emph{Security and Privacy, 2001. S\&P 2001. Proceedings. 2001 IEEE
  Symposium on}.\hskip 1em plus 0.5em minus 0.4em\relax IEEE, 2001, pp.
  130--143.

\bibitem{CUSUM}
E.~S. Page, ``Continuous inspection schemes,'' \emph{Biometrika}, vol.~41, no.
  1/2, pp. 100--115, 1954.

\bibitem{Moustakides}
G.~V. Moustakides \emph{et~al.}, ``Optimal stopping times for detecting changes
  in distributions,'' \emph{The Annals of Statistics}, vol.~14, no.~4, pp.
  1379--1387, 1986.

\bibitem{MEI}
Y.~Mei, ``Efficient scalable schemes for monitoring a large number of data
  streams,'' \emph{Biometrika}, vol.~97, no.~2, pp. 419--433, 2010.

\bibitem{taposh_hub}
T.~Banerjee and A.~O. Hero, ``Quickest hub discovery in correlation graphs,''
  in \emph{Signals, Systems and Computers, 2016 50th Asilomar Conference
  on}.\hskip 1em plus 0.5em minus 0.4em\relax IEEE, 2016, pp. 1248--1255.

\bibitem{GEM}
A.~O. Hero, ``Geometric entropy minimization (gem) for anomaly detection and
  localization,'' in \emph{Advances in Neural Information Processing Systems},
  2007, pp. 585--592.

\bibitem{GEM-2}
K.~Sricharan and A.~O. Hero, ``Efficient anomaly detection using bipartite k-nn
  graphs,'' in \emph{Advances in Neural Information Processing Systems}, 2011,
  pp. 478--486.

\bibitem{LMVS}
C.~D. Scott and R.~D. Nowak, ``Learning minimum volume sets,'' \emph{Journal of
  Machine Learning Research}, vol.~7, no. Apr, pp. 665--704, 2006.

\bibitem{zhao}
M.~Zhao and V.~Saligrama, ``Anomaly detection with score functions based on
  nearest neighbor graphs,'' in \emph{Advances in neural information processing
  systems}, 2009, pp. 2250--2258.

\bibitem{chen}
H.~Chen, ``Sequential change-point detection based on nearest neighbors,''
  \emph{arXiv preprint arXiv:1604.03611}, 2016.

\bibitem{zambon}
D.~Zambon, C.~Alippi, and L.~Livi, ``Concept drift and anomaly detection in
  graph streams,'' \emph{IEEE transactions on neural networks and learning
  systems}, no.~99, pp. 1--14, 2018.

\bibitem{lorden}
G.~Lorden \emph{et~al.}, ``Procedures for reacting to a change in
  distribution,'' \emph{The Annals of Mathematical Statistics}, vol.~42, no.~6,
  pp. 1897--1908, 1971.

\bibitem{odit}
Y.~Yilmaz, ``Online nonparametric anomaly detection based on geometric entropy
  minimization,'' in \emph{Information Theory (ISIT), 2017 IEEE International
  Symposium on}.\hskip 1em plus 0.5em minus 0.4em\relax IEEE, 2017, pp.
  3010--3014.

\bibitem{icmla}
Y.~Yilmaz and S.~Uludag, ``Mitigating iot-based cyberattacks on the smart
  grid,'' in \emph{Machine Learning and Applications (ICMLA), 2017 16th IEEE
  International Conference on}.\hskip 1em plus 0.5em minus 0.4em\relax IEEE,
  2017, pp. 517--522.

\bibitem{itsc}
A.~Haydari and Y.~Yilmaz, ``Real-time detection and mitigation of ddos attacks
  in intelligent transportation systems,'' in \emph{2018 21st International
  Conference on Intelligent Transportation Systems (ITSC)}.\hskip 1em plus
  0.5em minus 0.4em\relax IEEE, 2018, pp. 157--163.

\bibitem{agresti2018introduction}
A.~Agresti, \emph{An introduction to categorical data analysis}.\hskip 1em plus
  0.5em minus 0.4em\relax Wiley, 2018.

\bibitem{pval1}
M.~Baker, ``Statisticians issue warning over misuse of p values,'' \emph{Nature
  News}, vol. 531, no. 7593, p. 151, 2016.

\bibitem{pval2}
A.~Gelman, ``The problems with p-values are not just with p-values,'' \emph{The
  American Statistician}, vol.~70, 2016.

\bibitem{dtheory}
H.~Weyl, ``{\"U}ber die gleichverteilung von zahlen mod. eins,''
  \emph{Mathematische Annalen}, vol.~77, no.~3, pp. 313--352, 1916.

\bibitem{dnorm}
B.~A. Moser and T.~Natschl{\"a}ger, ``On stability of distance measures for
  event sequences induced by level-crossing sampling.'' \emph{IEEE Trans.
  Signal Processing}, vol.~62, no.~8, pp. 1987--1999, 2014.

\bibitem{Muja}
M.~Muja and D.~G. Lowe, ``Scalable nearest neighbor algorithms for high
  dimensional data,'' \emph{IEEE Transactions on Pattern Analysis \& Machine
  Intelligence}, no.~11, pp. 2227--2240, 2014.

\bibitem{kclass1}
K.~Fukunaga and J.~M. Mantock, ``A nonparametric two-dimensional display for
  classification,'' \emph{IEEE transactions on pattern analysis and machine
  intelligence}, no.~4, pp. 427--436, 1982.

\bibitem{kclass2}
J.~J. Remus, K.~D. Morton, P.~A. Torrione, S.~L. Tantum, and L.~M. Collins,
  ``Comparison of a distance-based likelihood ratio test and k-nearest neighbor
  classification methods,'' in \emph{Machine Learning for Signal Processing,
  2008. MLSP 2008. IEEE Workshop on}.\hskip 1em plus 0.5em minus 0.4em\relax
  IEEE, 2008, pp. 362--367.

\bibitem{DDoSDouligeris}
C.~Douligeris and A.~Mitrokotsa, ``Ddos attacks and defense mechanisms:
  classification and state-of-the-art,'' \emph{Computer Networks}, vol.~44,
  no.~5, pp. 643--666, 2004.

\bibitem{kolias2017ddos}
C.~Kolias, G.~Kambourakis, A.~Stavrou, and J.~Voas, ``Ddos in the iot: Mirai
  and other botnets,'' \emph{Computer}, vol.~50, no.~7, pp. 80--84, 2017.

\bibitem{IoTDataset}
Y.~Meidan, M.~Bohadana, Y.~Mathov, Y.~Mirsky, A.~Shabtai, D.~Breitenbacher, and
  Y.~Elovici, ``N-baiot—network-based detection of iot botnet attacks using
  deep autoencoders,'' \emph{IEEE Pervasive Computing}, vol.~17, no.~3, pp.
  12--22, 2018.

\bibitem{IoTDataset2}
Y.~Mirsky, T.~Doitshman, Y.~Elovici, and A.~Shabtai, ``Kitsune: an ensemble of
  autoencoders for online network intrusion detection,'' \emph{arXiv preprint
  arXiv:1802.09089}, 2018.

\end{thebibliography}


\begin{thebibliography}{9}

\bibitem{newinfobased}
Xiang, Yang, Ke Li, and Wanlei Zhou. "Low-rate DDoS attacks detection and traceback by using new information metrics." IEEE transactions on information forensics and security 6.2 (2011): 426-437.

\bibitem{medicAD}
Zhang, Haibin, Jiajia Liu, and Nei Kato. "Threshold tuning-based wearable sensor fault detection for reliable medical monitoring using Bayesian network model." IEEE Systems Journal 12.2 (2018): 1886-1896.

\bibitem{SURVEY}
Chandola, Varun, Arindam Banerjee, and Vipin Kumar. "Anomaly detection: A survey." ACM computing surveys (CSUR) 41.3 (2009): 15.

\bibitem{CUSUM}
E.S. Page, “Continuous inspection schemes”, Biometrika, vol.41, pp.100 - 115, 1954.

\bibitem{Moustakides}
G.V. Moustakides, "Optimal stopping times for detecting changes in distributions", The Annals of Statistics, vol. 14, no. 4, pp. 1379-1387, 1986.

\bibitem{MEI}
Mei, Y. (2010). Efficient scalable schemes for monitoring a large number of data streams. Biometrika 97, 2, 419 – 433.

\bibitem{taposh1}
Banerjee, Taposh, Hamed Firouzi, and Alfred O. Hero III. "Quickest detection for changes in maximal knn coherence of random matrices." arXiv preprint arXiv:1508.04720 (2015).

\bibitem{taposh_hub}
Banerjee, Taposh, and Alfred O. Hero. "Quickest hub discovery in correlation graphs." Signals, Systems and Computers, 2016 50th Asilomar Conference on. IEEE, 2016.

\bibitem{JIRAK}
Jirak, Moritz. "Uniform change point tests in high dimension." The Annals of Statistics 43.6 (2015): 2451-2483.

\bibitem{BIGDATASURVEY}
Habeeb, Riyaz Ahamed Ariyaluran, et al. "Real-time big data processing for anomaly detection: A Survey." International Journal of Information Management (2018).

\bibitem{CORR}
Avanesov, Valeriy, and Nazar Buzun. "Change-point detection in high-dimensional covariance structure." arXiv preprint arXiv:1610.03783 (2016).

\bibitem{GEM} 
Hero, Alfred O. "Geometric entropy minimization (GEM) for anomaly detection and localization." Advances in Neural Information Processing Systems. 2007.

\bibitem{GEM-2}
Sricharan, Kumar, and Alfred O. Hero. "Efficient anomaly detection using bipartite k-NN graphs." Advances in Neural Information Processing Systems. 2011.

\bibitem{chen}
Chen, Hao. "Sequential change-point detection based on nearest neighbors." arXiv preprint arXiv:1604.03611 (2016).

\bibitem{zambon}
Zambon, Daniele, Cesare Alippi, and Lorenzo Livi. "Concept drift and anomaly detection in graph streams." IEEE transactions on neural networks and learning systems 99 (2018): 1-14.

\bibitem{lorden}
Lorden, Gary. "Procedures for reacting to a change in distribution." The Annals of Mathematical Statistics 42.6 (1971): 1897-1908.

\bibitem{zhao}
Zhao, Manqi, and Venkatesh Saligrama. "Anomaly detection with score functions based on nearest neighbor graphs." Advances in neural information processing systems. 2009.

\bibitem{LMVS}
Scott, Clayton D., and Robert D. Nowak. "Learning minimum volume sets." Journal of Machine Learning Research 7.Apr (2006): 665-704.

\bibitem{svmbased}
Steinwart, Ingo, Don Hush, and Clint Scovel. "A classification framework for anomaly detection." Journal of Machine Learning Research 6.Feb (2005): 211-232.

\bibitem{infobased}
Lee, Wenke, and Dong Xiang. "Information-theoretic measures for anomaly detection." Security and Privacy, 2001. S\&P 2001. Proceedings. 2001 IEEE Symposium on. IEEE, 2001.

\bibitem{Muja}
Muja, Marius, and David G. Lowe. "Scalable nearest neighbor algorithms for high dimensional data." IEEE Transactions on Pattern Analysis \& Machine Intelligence 11 (2014): 2227-2240.

\bibitem{spacecraft}
Hundman, Kyle, et al. "Detecting Spacecraft Anomalies Using LSTMs and Nonparametric Dynamic Thresholding." arXiv preprint arXiv:1802.04431 (2018).

\bibitem{odit}
Yilmaz, Yasin, ``Online nonparametric anomaly detection based on geometric entropy minimization." \emph{2017 IEEE International Symposium on Information Theory (ISIT)}, pp. 3010-3014, 2017.

\bibitem{icmla}
Yilmaz, Yasin, and Uludag, Suleyman, ``Mitigating IoT-based Cyberattacks on the Smart Grid", \emph{IEEE International Conference on Machine Learning and Applications (ICMLA)}, 2017.

\bibitem{itsc}
Haydari, Ammar, and Yilmaz, Yasin, ``Real-Time Detection and Mitigation of DDoS Attacks in Intelligent Transportation Systems", \emph{IEEE International Conference on Intelligent Transportation Systems (ITSC)}, 2018.

\bibitem{pval1}
Baker, Monya, ``Statisticians issue warning on p values," \emph{Nature}, vol. 531, no. 7593, pp. 151, 2016.

\bibitem{pval2}
Gelman, Andrew, ``The problems with p-values are not just with p-values," \emph{The American Statistician}, 2016.

\bibitem{agresti2018introduction}
A. Agresti, \emph{An introduction to categorical data analysis}, Wiley, 2018.

\bibitem{dtheory}
H. Weyl, ``Uber die Gleichverteilung von Zahlen mod. Eins", \emph{Math. Ann.}, vol. 77, pp. 313?352, 1916.

\bibitem{dnorm}
B. A. Moser and T. Natschlager, ``On stability of distance measures for event sequences induced by level-crossing sampling", \emph{IEEE Trans. Signal Process.}, vol. 62, no. 8, pp. 1987?1999, 2014.

\bibitem{N-BaIoT}
Meidan, Yair, et al. "N-BaIoT: Network-based Detection of IoT Botnet Attacks Using Deep Autoencoders." arXiv preprint arXiv:1805.03409 (2018).

\bibitem{DDoSTechniques}
Carl, Glenn, et al. "Denial-of-service attack-detection techniques." IEEE Internet computing 10.1 (2006): 82-89.

\bibitem{DDoSAttacks}
Lau, Felix, et al. "Distributed denial of service attacks." Systems, Man, and Cybernetics, 2000 IEEE International Conference on. Vol. 3. IEEE, 2000.

\bibitem{DDoSDouligeris}
Douligeris, Christos, and Aikaterini Mitrokotsa. "DDoS attacks and defense mechanisms: classification and state-of-the-art." Computer Networks 44.5 (2004): 643-666.

\bibitem{IoT_DDoS}
Kolias, Constantinos, et al. "DDoS in the IoT: Mirai and other botnets." Computer 50.7 (2017): 80-84.

\bibitem{DDoS_on_IoT}
Sonar, Krushang, and Hardik Upadhyay. "A survey: DDOS attack on Internet of Things." International Journal of Engineering Research and Development 10.11 (2014): 58-63.

\bibitem{IoTDataset}
Meidan, Yair, et al. "N-BaIoT—Network-Based Detection of IoT Botnet Attacks Using Deep Autoencoders." IEEE Pervasive Computing 17.3 (2018): 12-22.

\bibitem{IoTDataset2}
Mirsky, Yisroel, et al. "Kitsune: an ensemble of autoencoders for online network intrusion detection." arXiv preprint arXiv:1802.09089 (2018).



\bibitem{CTUDataset}
Garcia, Sebastian, et al. "An empirical comparison of botnet detection methods." computers \& security 45 (2014): 100-123.

\end{thebibliography}

\ignore{

}

\end{document}